# Exploiting deep residual networks for human action recognition from skeletal data

Huy-Hieu Pham[a,b,**], Louahdi Khoudour[a], Alain Crouzil[b], Pablo Zegers[c], Sergio A. Velastin[d,e]

[a]*Centre d'Études et d'Expertise sur les Risques, l'Environnement, la Mobilité et l'Aménagement (CEREMA), 31400 Toulouse, France*
[b]*Institut de Recherche en Informatique de Toulouse (IRIT), Université de Toulouse, UPS, 31062 Toulouse Cedex 9, France*
[c]*Aparnix, La Gioconda 4355, 10B, Las Condes, Santiago, Chile*
[d]*Department of Computer Science, Applied Artificial Intelligence Research Group, University Carlos III de Madrid, 28270 Madrid, Spain*
[e]*School of Electronic Engineering and Computer Science, Queen Mary University of London, UK*

ABSTRACT

The computer vision community is currently focusing on solving action recognition problems in real videos, which contain thousands of samples with many challenges. In this process, Deep Convolutional Neural Networks (D-CNNs) have played a significant role in advancing the state-of-the-art in various vision-based action recognition systems. Recently, the introduction of residual connections in conjunction with a more traditional CNN model in a single architecture called Residual Network (ResNet) has shown impressive performance and great potential for image recognition tasks. In this paper, we investigate and apply deep ResNets for human action recognition using skeletal data provided by depth sensors. Firstly, the 3D coordinates of the human body joints carried in skeleton sequences are transformed into image-based representations and stored as RGB images. These color images are able to capture the spatial-temporal evolutions of 3D motions from skeleton sequences and can be efficiently learned by D-CNNs. We then propose a novel deep learning architecture based on ResNets to learn features from obtained color-based representations and classify them into action classes. The proposed method is evaluated on three challenging benchmark datasets including MSR Action 3D, KARD, and NTU-RGB+D datasets. Experimental results demonstrate that our method achieves state-of-the-art performance for all these benchmarks whilst requiring less computation resource. In particular, the proposed method surpasses previous approaches by a significant margin of **3.4**% on MSR Action 3D dataset, **0.67**% on KARD dataset, and **2.5**% on NTU-RGB+D dataset.

## 1. Introduction

Human Action Recognition (HAR) is one of the key fields in computer vision and plays an important role in many intelligent systems involving video surveillance, human-machine interaction, self-driving cars, robot vision and so on. The main goal of this field is to determine, and then recognize what humans do in unknown videos. Although significant progress has been made in the last years, accurate action recognition in videos is still a challenging task due to many obstacles such as viewpoint, occlusion or lighting conditions (Poppe, 2010).

Traditional studies on HAR mainly focus on the use of hand-crafted local features such as Cuboids (Dollár et al., 2005) or HOG/HOF (Laptev et al., 2008) that are provided by 2D cameras. These approaches typically recognize human actions based on the appearance and movements of human body parts in videos. Another approach is to use **G**enetic **P**rogramming (**GP**) for generating spatio-temporal descriptors of motions (Liu et al., 2012). However, one of the major limitations of the 2D data is the absence of 3D structure from the scene. There-fore, single modality action recognition on RGB sequences is not enough to overcome the challenges in HAR, especially in realistic videos. Recently, the rapid development of depth-sensing time-of-flight camera technology has helped in dealing with problems, which are considered complex for traditional cameras. Depth cameras, *e.g.*, Microsoft Kinect TMsensor (Cruz et al., 2012; Han et al., 2013) or ASUS Xtion (ASUS, 2018), are able to provide detailed information about the 3D structure of the human motion. Thus, many approaches have been proposed for recognizing actions based on RGB sequences, depth (Baek et al., 2017), or combining these two data types (RGB-D) (Wang et al., 2014), which are provided by depth sensors. Moreover, they are also able to provide real-time skeleton estimation algorithms (Shotton et al., 2013) that help to describe actions in a more precise and effective way. The skeleton-based representations have the advantage of lower dimensionality than RGB/RGB-D-based representations. This benefit makes action recognition systems become simpler and faster. Therefore, exploiting the 3D skeletal data provided by depth sensors for HAR is a promising research direction. In fact, many skeleton-based action recognition approaches have been proposed (Wang et al., 2012; Xia et al., 2012b; Chaudhry et al., 2013; Vemulapalli et al., 2014; Ding et al., 2016).

**Corresponding author: Tel.: +33-605-568-269;
*e-mail:* huy-hieu.pham@cerema.fr (Huy-Hieu Pham)

In recent years, approaches based on Convolutional Neural Networks (CNNs) have achieved outstanding results in many image recognition tasks (Krizhevsky et al., 2012; Karpathy et al., 2014). After the success of AlexNet (Krizhevsky et al., 2012) in the ImageNet competition (Russakovsky et al., 2015), a new direction of research has been opened for finding higher performing CNN architectures. As a result, there are many signs that seem to indicate that the learning performance of CNNs can be significantly improved by increasing their depth (Simonyan and Zisserman, 2014b; Szegedy et al., 2015; Telgarsky, 2016). In the literature of HAR, many studies have indicated that CNNs have the ability to learn complex motion features better than hand-crafted approaches (see Figure 1). However, most authors have just focused on the use of relatively small and simple CNNs such as AlexNet (Krizhevsky et al., 2012) and have not yet fully exploited the potential of recent state-of-the-art very deep CNN architectures. In addition, most existing CNN-based approaches use RGB, depth or RGB-D sequences as the input to learning models. Although RGB-D images are informative for action recognition, however, the computation complexity of these models will increases rapidly when the dimension of the input features is large. This makes models become more complex, slower and less practical for solving large-scale problems as well as real-time applications.

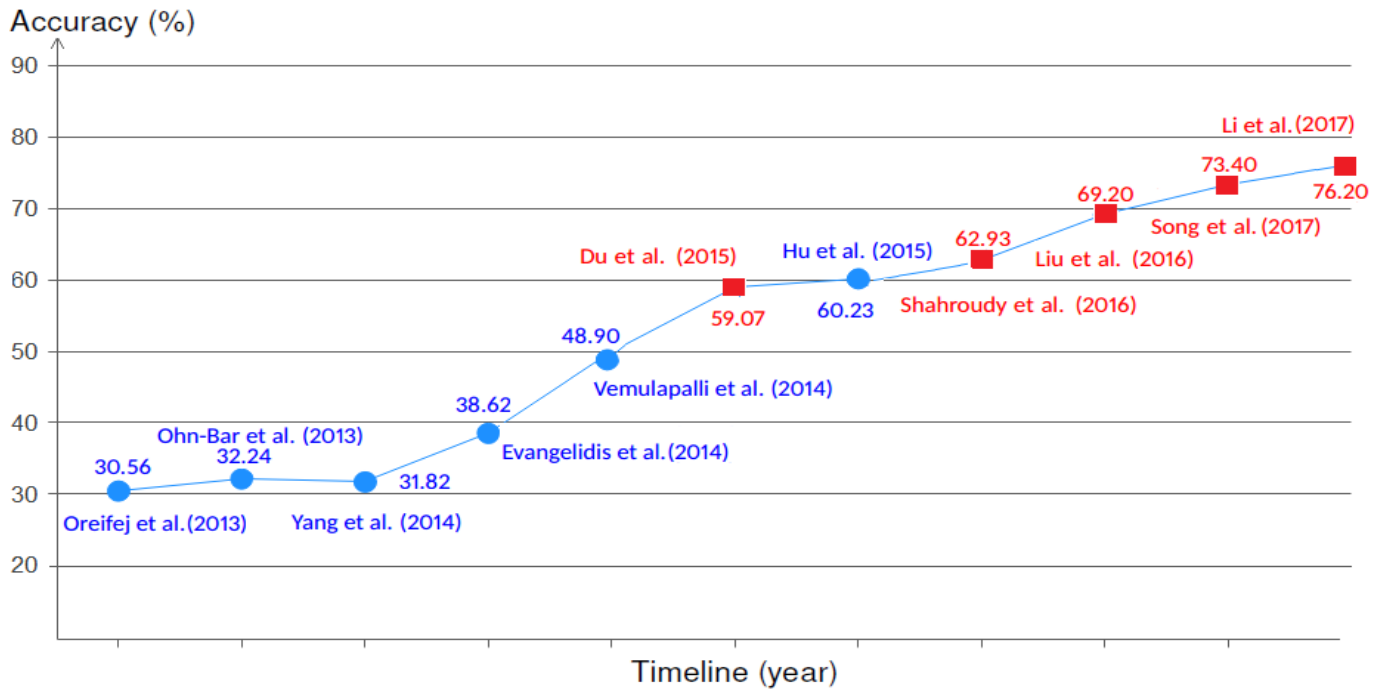

**Fig. 1. The recognition performance of hand-crafted and deep learning approaches reported on the Cross-View evaluation criteria of NTU-RGB+D dataset (Shahroudy et al., 2016). The traditional approaches are marked with circles (Oreifej and Liu, 2013; Ohn-Bar and Trivedi, 2013; Yang and Tian, 2014; Evangelidis et al., 2014; Hu et al., 2015). The deep learning based approaches are marked with squares (Du et al., 2015; Shahroudy et al., 2016; Liu et al., 2016; Song et al., 2017; Li et al., 2017).**

In this paper, we aim to take full advantages of 3D skeleton-based representations and the ability of learning highly hierarchical image features of Deep Convolutional Neural Networks (D-CNNs) to build an end-to-end learning framework for HAR from skeletal data. To this end, all the 3D coordinates of the skeletal joints in the body provided by Kinect sensors are represented as 3D arrays and then stored as RGB images by using a simple skeleton-to-image encoding method. The main goal of this processing step is to ensure that the color images effectively represents the spatio-temporal structure of the human action carried in skeleton sequences and they are compatible by the deep learning networks as D-CNNs. To learn image features and recognize their labels, we propose to use Residual Networks (ResNets) (He et al., 2016) – a very deep and recent state-of-the-art CNN for image recognition. In the hope of achieving higher levels of performance, we propose a novel deep architecture based on the original ResNets, which is easier to optimize and able to prevent overfitting better. We evaluate the proposed method on three benchmark skeleton datasets (MSR Action 3D (Li et al., 2010); Kinect Activity Recognition Dataset - KARD (Gaglio et al., 2015); NTU-RGB+D (Shahroudy et al., 2016)) and obtain state-of-the-art recognition accuracies on all these datasets. Furthermore, we also point out the effectiveness of our learning framework in terms of computational complexity, the ability to prevent overfitting and to reduce the effect of degradation phenomenon in training very deep networks.

The contributions of our work lie in the following aspects:

- Firstly, we propose an end-to-end learning framework based on ResNets to effectively learn the spatial-temporal evolutions carried in RGB images which encoded from skeleton sequences for 3D human action recognition. To the best of our knowledge, this is the first time ResNet-based models are applied successfully on skeletal data to recognize human actions.
- Secondly, we present a novel ResNet building unit to construct very deep ResNets. Our experiments on action recognition tasks prove that the proposed architecture is able to learn features better than the original ResNet model (He et al., 2016). This architecture is general and could be applied for various image recognition problems, not only the human action recognition.
- Finally, we show the effectiveness of our learning framework on action recognition tasks by achieving the state-of-the-art performance on three benchmark datasets including the most challenging skeleton benchmark currently available, whilst requiring less computation.

The rest of the paper is organized as follows: Section 2 discusses related works. In section 3, we present the details of our proposed method. Datasets and experiments are described in Section 4. Experimental results are shown in Section 5. In Section 6, we discuss classification accuracy, overfitting issues, degradation phenomenon and computational efficiency of the proposed deep learning networks. This section will also discuss about different factors that affect the recognition rate. Finally, Section 7 concludes the paper and discusses our future work.

## 2. Related Work

Our study is closely related to two major topics: skeleton-based action recognition and designing D-CNN architectures for visual recognition tasks. This section presents some key studies on these topics. We first discuss previous works on skeleton-based action recognition. Then, we introduce an overview of the development of D-CNNs and their potential for HAR. Related to HAR based on RGB/RGB-D sequences, we refer the interested reader to the most successful approaches including Bag of Words (BoWs) (Peng et al., 2016; Liu et al., 2017a; Wang and Schmid, 2013), Dynamic Image Networks (Bilen et al., 2016) and D-CNNs to learn RGB representation from raw data (Ng et al., 2015; Simonyan and Zisserman, 2014a).

***Skeleton-based action recognition***: The 3D skeletal data provided by depth sensors has been extensively exploited for HAR. Recent skeleton-based action recognition methods can be divided into two main groups. The first group combines hand-crafted skeleton features and graphical models to recognize actions. The spatio-temporal representations from skeleton sequences are often modeled by several common probabilistic graphical models such as **H**idden **M**arkov **M**odel (**HMM**) (Lv and Nevatia, 2006; Wang et al., 2012; Yang et al., 2013), **L**atent **D**irichlet **A**llocation (**LDA**) (Blei et al., 2003; Liu et al., 2012) or **C**onditional **R**andom **F**ield (**CRF**) (Koppula and Saxena, 2013). In addition, **F**ourier **T**emporal **P**yramid (**FTP**) (Wang et al., 2012; Vemulapalli et al., 2014; Hu et al., 2015) has also been used to capture the temporal dynamics of actions and then to predict their labels. Another solution based on shape analysis methods has been exploited for skeleton-based human action (Amor et al., 2016; Devanne et al., 2013). Specifically, the authors defined an action as a sequence of skeletal shapes and analyzed them by a statistical shape analysis tool such as the geometry of Kendall's shape manifold. Typical classifiers, *e.g.*, **K**-**N**earest-**N**eighbor (**KNN**) or **S**upport **V**ector **M**achine (**SVM**) were then used for classification. Although promising results have been achieved, however, most of these works require a lot of feature engineering. *E.g.*, the skeleton sequences often need to be segmented and aligned for HMM- and CRF-based approaches. Meanwhile, FTP-based approaches cannot globally capture the temporal sequences of actions.

The second group of methods is based on **R**ecurrent **N**eural **N**etworks with **L**ong **S**hort-**T**erm **M**emory **N**etwork (**RNN-LSTMs**) (Hochreiter and Schmidhuber, 1997). The architecture of an RNN-LSTM network allows to store and access the long range contextual information of a temporal sequence. As human skeleton-based action recognition can be regarded as a time-series problem (Gong et al., 2014), RNN-LSTMs can be used to learn human motion features from skeletal data. For that reason, many authors have explored RNN-LSTMs for 3D HAR from skeleton sequences (Du et al., 2015; Veeriah et al., 2015; Ling et al., 2016; Zhu et al., 2016; Shahroudy et al., 2016; Liu et al., 2016). To better capture the spatio-temporal dynamics carried in skeletons, some authors used a CNN as a visual feature extractor, combined with a RNN-LSTM in a unified framework for modeling human motions (Mahasseni and Todorovic, 2016; Shi and Kim, 2017; Kim and Reiter, 2017). Although RNN-LSTM-based approaches have been reported to provide good performance. However, there are some limitations that are difficult to overcome. *I.e.*, the use of RNNs can lead to overfitting problems when the number of input features is not enough for training network. Meanwhile, the computational time can become a serious problem when the input features increase.

***D-CNNs for visual recognition***: CNNs have led to a series of breakthroughs in image recognition and related tasks. Recently, there is growing evidence that D-CNN models can improve performance in image recognition (Simonyan and Zisserman, 2014b; Szegedy et al., 2015). However, deep networks are very difficult to train. Two main reasons that impede the convergence of deeper networks are vanishing gradients problems (Glorot and Bengio, 2010) and degradation phenomenon (He and Sun, 2015). The vanishing gradients problem occurs when the network is deep enough, the error signal from the output layer can be completely attenuated on its way back to the input layer. This obstacle has been solved by normalized initialization (LeCun et al., 1998; He et al., 2015), especially by using Batch Normalization (Ioffe and Szegedy, 2015a). When the deep networks start converging, a degradation phenomenon occurs (see an example in Figure 2). If we add more layers to a deep network, this can lead to higher training and/or testing error (He and Sun, 2015). This phenomenon is not as simple as an overfitting problem. To reduce the effect of vanishing gradients problems and degradation phenomenon, (He et al., 2016) introduced Residual Networks (ResNets) with the presence of shortcut connections parallel to their traditional convolutional layers. This idea helps ResNets to improve the information flow across layers. Experimental results on two well-known datasets including CIFAR-10 (Krizhevsky, 2009) and ImageNet (Russakovsky et al., 2015) confirmed that ResNets can improve the recognition performance and reduce degradation phenomenon.

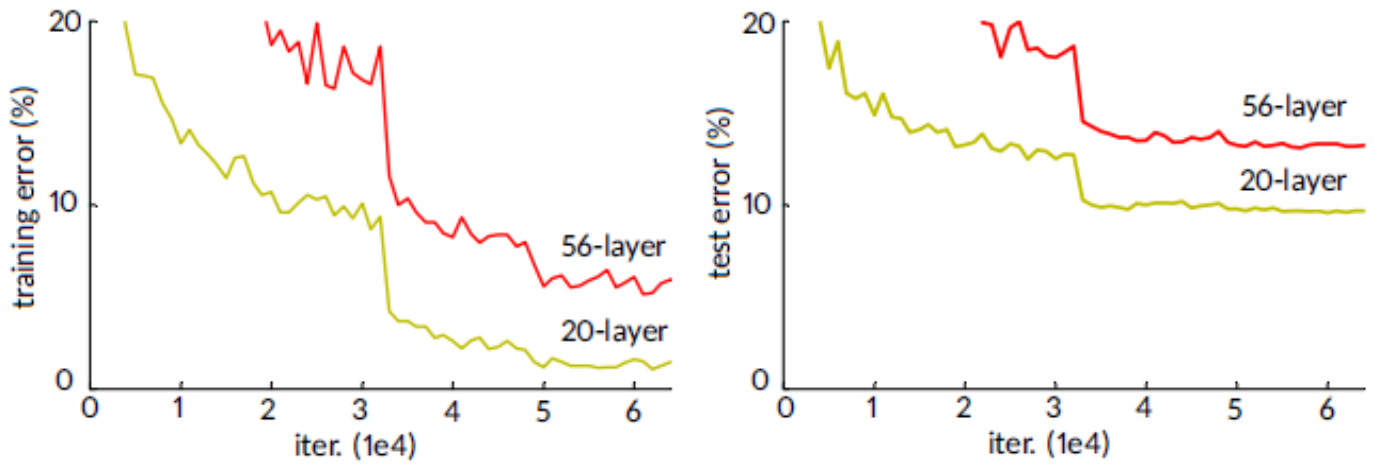

(a) (b)

**Fig. 2. (a) Training error and (b) test error on CIFAR-10 (Krizhevsky, 2009) with 20-layer and 56-layer CNNs reported by (He et al., 2016). The deeper network has higher error for both training and test phases.**

Several authors have exploited the feature learning ability of CNNs on skeletal data (Hou et al., 2017; Wang et al., 2016; Song et al., 2017; Li et al., 2017). However, such studies mainly focus on finding good skeletal representations and learning features with simple CNN architectures. In contrast, in this paper we concentrate on exploiting the power of D-CNNs for action recognition using a simple skeleton-based representation. We investigate and design a novel deep learning framework based on ResNet (He et al., 2016) to learn action features from skeleton sequences and then classify them into classes. Our exper-imental results show state-of-the-art performance on the MSR Action 3D (Li et al., 2010), KARD (Gaglio et al., 2015) and NTU-RGB+D (Shahroudy et al., 2016) datasets. Besides, our proposed solution is general and can be applied on various different types of input data. For instance, this idea could be applied on the motion capture (MoCap) data provided by inertial sensors.

## 3. Method

This section presents our proposed method. We first describe a technique allowing to encode the spatio-temporal information of skeleton sequences into RGB images. Then, a novel ResNet architecture is proposed for learning and recognizing actions from obtained RGB images. Before that, in order to put our method into context, it is useful to review the central ideas behind the original ResNet (He et al., 2016) architecture.

### *3.1. Encoding skeletal data into RGB images*

Currently, the real-time skeleton estimation algorithms have been integrated into commercial depth cameras (Shotton et al., 2013). This technology allows to quickly and easily extract the position of the joints in the human body (Figure 3), which is suitable for the problem of 3D action recognition. One of the major challenges in exploiting CNN-based methods for skeleton-based action recognition is how a temporal skeleton sequence can be effectively represented and fed to CNNs for learning data features and perform classification. As CNNs are able to work well on still images, the idea therefore is to encode the spatial and temporal dynamics of skeleton sequences into 2D image structures. In general, two essential elements for recognizing actions are static postures and their temporal dynamics. These two elements can be transformed into the static spatial structure of a color image (Bilen et al., 2016; Hou et al., 2017; Wang et al., 2016). Then, a representation learning model such as CNNs can be deployed to learn image features and classify them into classes in order to recognize the original skeleton sequences.

Given a skeleton sequence $\mathcal{S}$ with N frames, denoting as $\mathcal{S}$ $= \{F_1, F_2, ..., F_N\}$. To represent the spatio-temporal information of a skeleton sequence as an RGB image, we transform the 3D joint coordinates $(x_i, y_i, z_i)$ carried in each skeleton $\{F_n\}$, $n \in [1, N]$ into the range of $[\mathbf{0}, \mathbf{255}]$ by normalizing these coordinates via a transformation function $\mathbf{F}(\cdot)$ as follows:

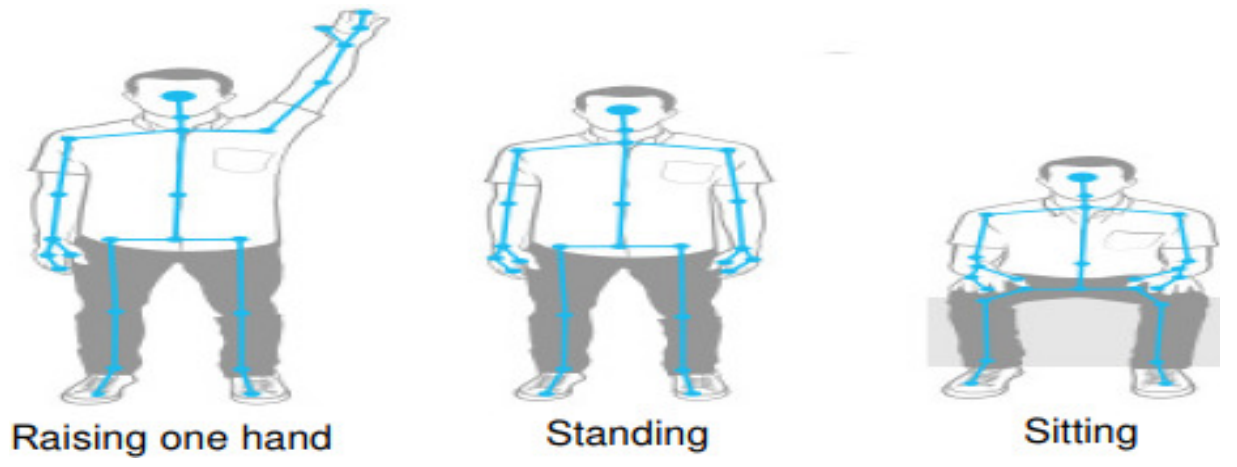

**Fig. 3. Illustration of the joint positions in the human body extracted by Kinect v2 sensor (Microsoft, 2014). A sequence of skeletons is able to describe correctly what a person performs in unseen videos.**

$$(x'_i, y'_i, z'_i) = \mathbf{F}(x_i, y_i, z_i) \tag{1}$$

$$x'_i = 255 \times \frac{(x_i - \min\{C\})}{\max\{C\} - \min\{C\}} \tag{2}$$

$$y'_i = 255 \times \frac{(y_i - \min\{C\})}{\max\{C\} - \min\{C\}} \tag{3}$$

$$z'_i = 255 \times \frac{(z_i - \min\{C\})}{\max\{C\} - \min\{C\}} \tag{4}$$

where $\min\{C\}$ and $\max\{C\}$ are the maximum and minimum values of all coordinates, respectively. The new coordinate space is quantified to integral image representation and three coordinates $(x'_i, y'_i, z'_i)$ are considered as the three components R,G,B of a color-pixel ($x'_i$ = R; $y'_i$ = G; $z'_i$ = B). As shown in Figure 4, each skeleton sequence is encoded into an RGB image. By this transformation, the raw data of skeleton sequences are converted to 3D tensors, which will then be fed into the learning model as the input features.

The order of joints in each frame is non-homogeneous for many skeleton datasets. Thus, it is necessary to rearrange joints and find a better representations in which different actions can be easily distinguished by the learning model. In other words, the image-based representation needs to contain discriminative

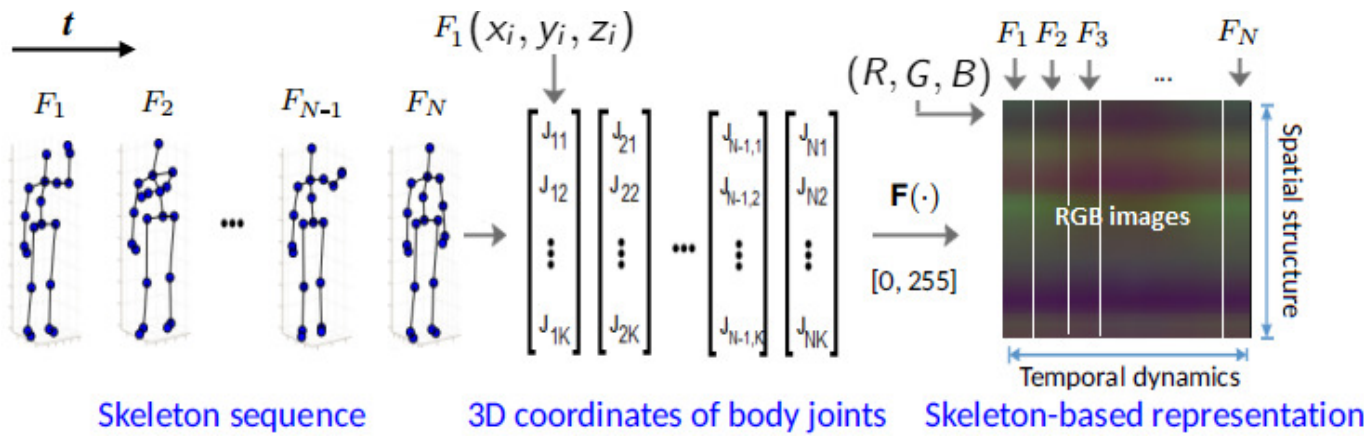

**Fig. 4. Illustration of the color encoding process. Here, $N$ denotes the number of frames in each skeleton sequence. $K$ denotes the number of joints in each frame. The value of $K$ depends on the depth sensors and data acquisition settings.**

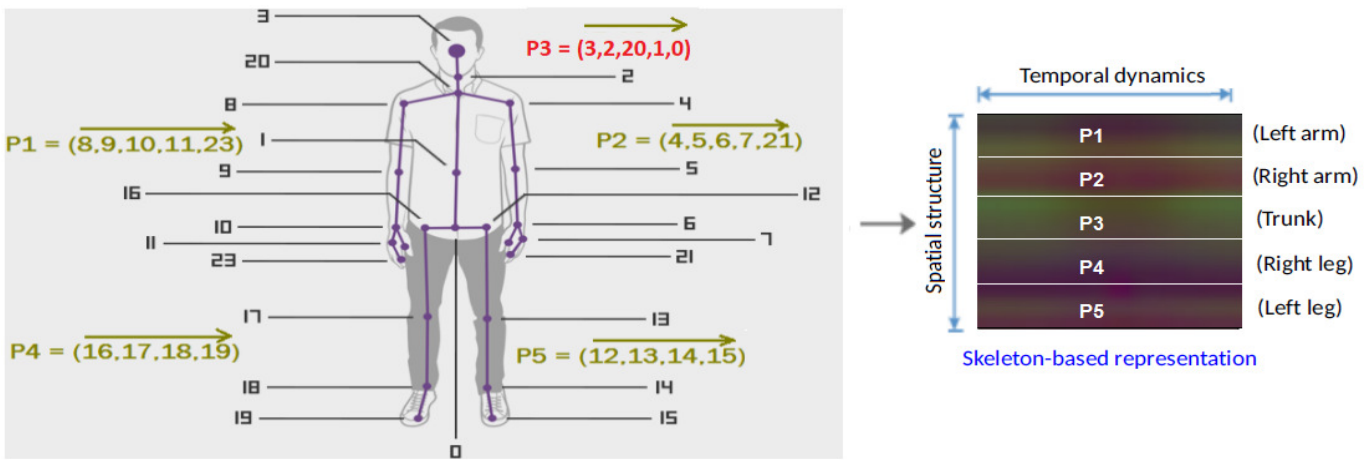

**Fig. 5. Arranging pixels in RGB images according to human body physical structure.**

features – a key factor to ensure the success of the CNNs during learning process. Naturally, the human body is structured by four limbs and one trunk. Simple actions can be performed through the movements of a limb while more complex actions come from the movements of a group of limbs in conjunction with the whole body. Inspired by this idea, (Du et al., 2015) proposed a simple and effective technique for representing skeleton sequences according to human body physical structure. To keep the local motion characteristics and to generate more discriminative features in image-based representations, we divide each skeleton frame into five parts, including two arms (**P1**, **P2**), two legs (**P4**, **P5**), and one trunk (**P3**). In each part from **P1** to **P5**, the joints are concatenated according to their physical connections. We then rearrange these parts in a sequential order, *i.e.*, **P1** → **P2** → **P3** → **P4** → **P5**. The whole process of rearranging all frames in a sequence can be done by rearranging the order of the rows of pixels in RGB-based representations as illustrated in Figure 5. Some skeleton-based representations obtained from the MSR Action 3D dataset (Li et al., 2010) are shown in Figure 6.

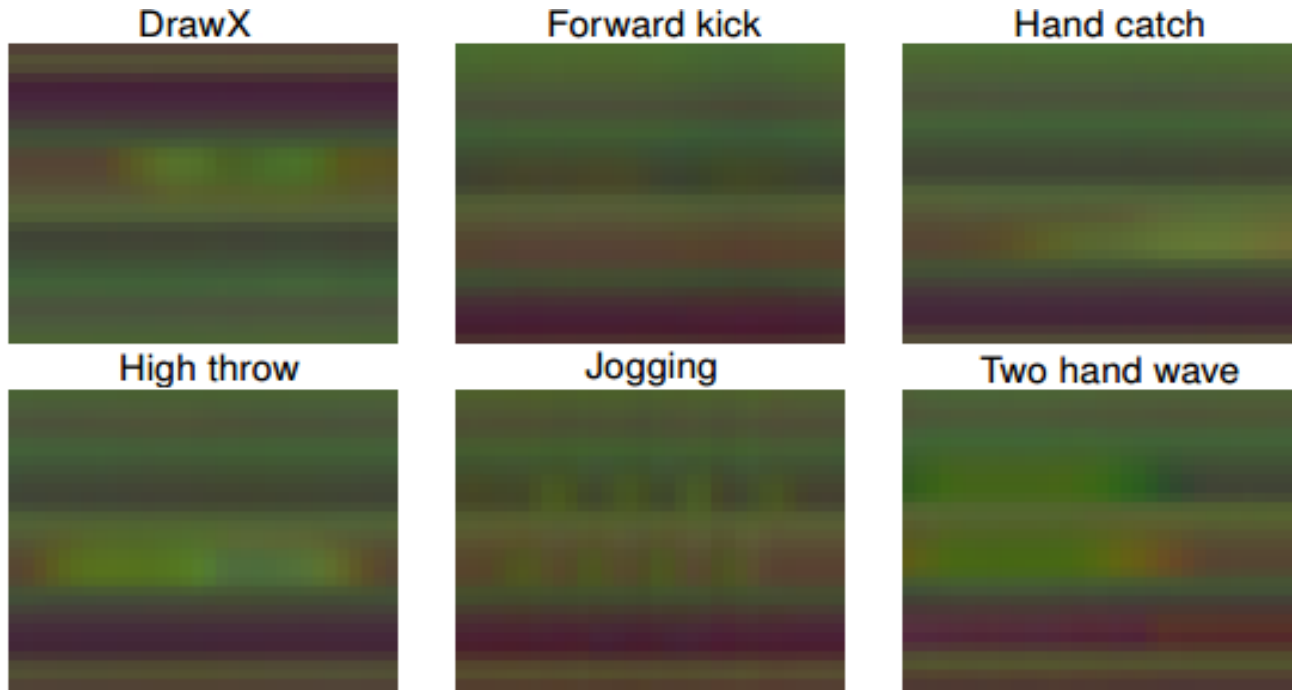

**Fig. 6. Output of the encoding process obtained from some samples of the MSR Action 3D dataset (Li et al., 2010). In our experiments, all images were resized to 32 × 32 pixels before feeding into the deep learning networks. Best viewed in color.**

### 3.2. Deep residual network

A simple difference between ResNets and traditional CNNs is that ResNets provide a clear path for gradients to back propagate to early layers during training. A deep ResNet is a modularized architecture that is constructed from multiple ResNet building units. Each unit has shortcut connection in parallel with traditional convolutional layers, which connects the input feature directly to its output. Considering the input feature of the $l^{th}$ layer as $x_l$, traditional CNNs (Figure 7a) learn a mapping function: $x_{l+1} = \mathcal{F}(x_l)$ where $x_{l+1}$ is the output of the $l^{th}$ layer, $\mathcal{F}(\cdot)$ is a non-linear transformation that can be implemented by the combination of **B**atch **N**ormalization (**BN**) (Ioffe and Szegedy, 2015b), **R**ectified **L**inear **U**nits (**ReLU**) (Nair and Hinton, 2010) and Convolutions. Different from traditional CNNs, a ResNet building unit (Figure 7b) performs the following computations:

$$x_{l+1} = \mathrm{ReLU}\left(\mathcal{F}(x_l, \mathcal{W}_l) + id(x_l)\right) \tag{5}$$

where $x_l$ and $x_{l+1}$ are input and output features of the $l^{th}$ ResNet unit, respectively; $id(x)$ is the identity function $id(x_l) = x_l$ and $\mathcal{W}_l$ is a set of weights and biases associated with the $l^{th}$ ResNet unit. The detailed architecture of an original ResNet unit is shown in Figure 8a. In this architecture, $\mathcal{F}(\cdot)$ consists of a series of layers: **Convolution-BN-ReLU-Convolution-BN**. The ReLU (Nair and Hinton, 2010) is applied after each element-wise addition ⊕.

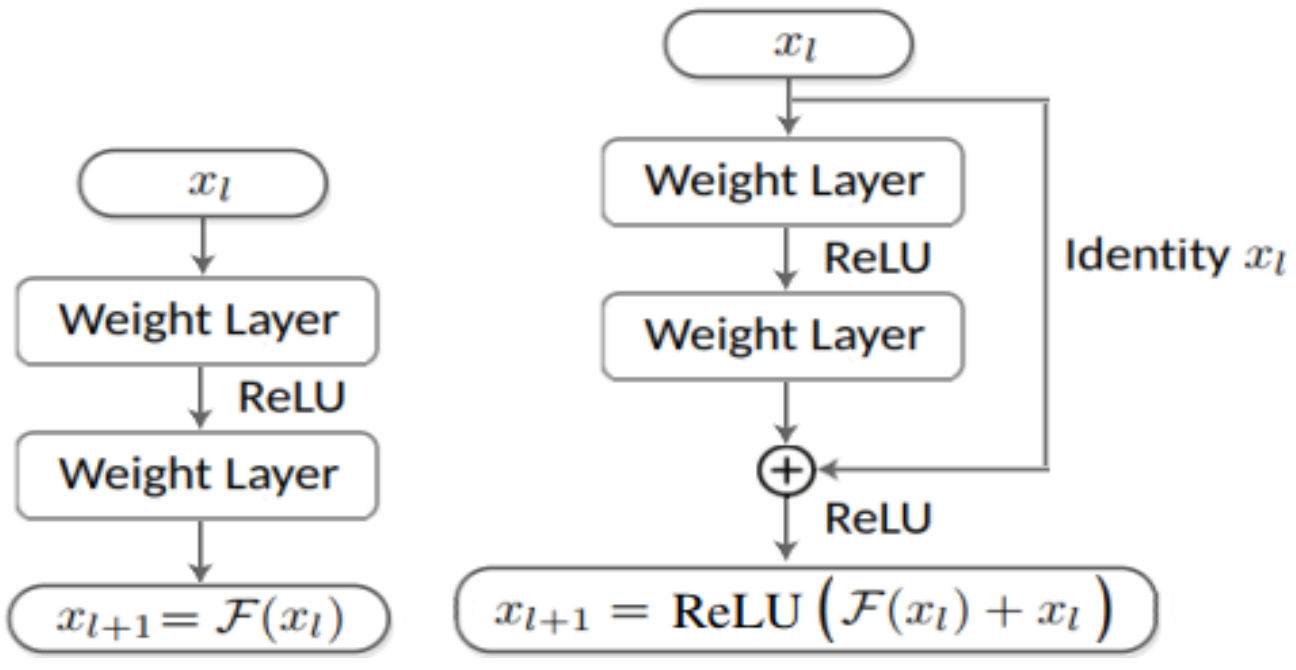

(a) (b)

**Fig. 7. (a) Information flow executed by a traditional CNN; (b) Information flow executed by a ResNet building unit (He et al., 2016).**

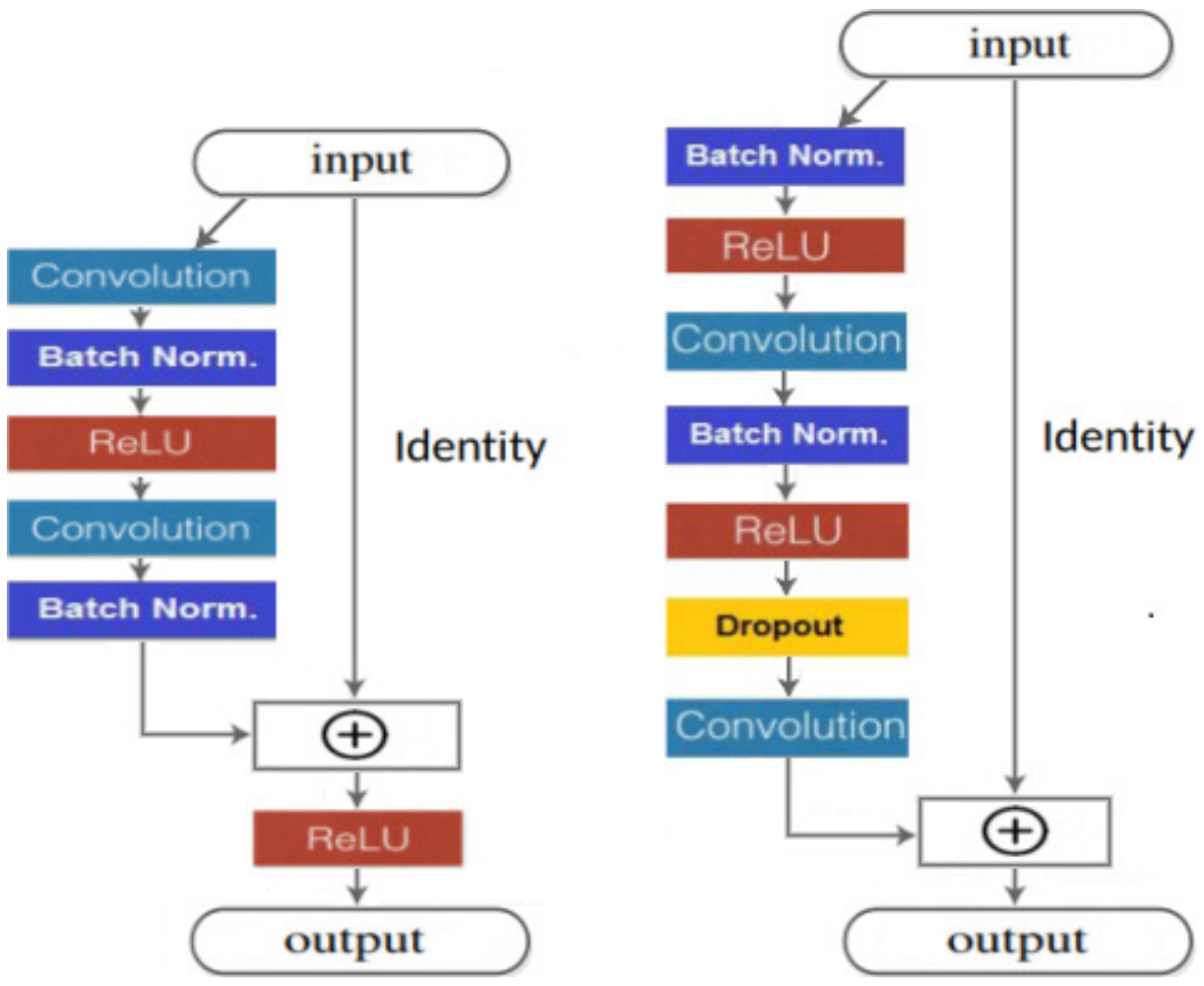

(a) (b)

**Fig. 8. (a) A ResNet building unit that was proposed in the original paper (He et al., 2016); (b) Our proposed ResNet building. The symbol ⊕ denotes element-wise addition.**

### *3.3. An improved residual network for skeleton-based action recognition*

The original ResNet architecture has a direct path for propagating information within a residual unit. However, the presence of non-linear activations as ReLUs (Nair and Hinton, 2010) behind element-wise additions ⊕ (see Figure 8a) means that the signal cannot be directly propagated from one block to any other block. To solve this problem, we propose an improved ResNet building block in which the signal can be directly propagated from any unit to another, both forward and backward for the entire network. The idea is to replace ReLU layers after each element-wise addition ⊕ by identity mappings $id(\cdot)$ for all units. That way, the information flow across each new ResNet unit can be rewritten as:

$$x_{l+1} = id(y_l) = \mathcal{F}(x_l, \mathcal{W}_l) + x_l \tag{6}$$

Eqn. (6) suggests that the feature $x_L$ of any deeper unit $L$ can be represented according to the feature $x_l$ of any shallower unit $l$:

$$x_L = x_l + \sum_{i=l}^{L-1} \mathcal{F}(x_i, \mathcal{W}_i) \tag{7}$$

Also, the feature $x_L$ can be represented according to the input feature $x_0$ of the first ResNet unit:

$$x_L = x_0 + \sum_{i=0}^{L-1} \mathcal{F}(x_i, \mathcal{W}_i) \tag{8}$$

Eqn. (8) indicates that we have created a **direct path** that helps the signal to be directly propagated in forward pass through the entire network. Considering now the backpropagation information, let $\mathcal{L}$ be the loss function that the network needs to optimize during the supervised training stage. From the chain rule of backpropagation (LeCun et al., 1989) and Eqn. (7), we can express the backpropagation information through layers as:

$$\frac{\partial \mathcal{L}}{\partial x_l} = \frac{\partial \mathcal{L}}{\partial x_L}\frac{\partial x_L}{\partial x_l} = \frac{\partial \mathcal{L}}{\partial x_L}\frac{\partial \left(x_l + \sum_{i=l}^{L-1} \mathcal{F}(x_i, \mathcal{W}_i)\right)}{\partial x_l} \tag{9}$$

or:

$$\frac{\partial \mathcal{L}}{\partial x_l} = \frac{\partial \mathcal{L}}{\partial x_L}\left(1 + \frac{\partial}{\partial x_l}\sum_{i=l}^{L-1} \mathcal{F}(x_i, \mathcal{W}_i)\right) \tag{10}$$

In Eqn. (10), the gradient $\frac{\partial \mathcal{L}}{\partial x_l}$ depends on two elements $\frac{\partial \mathcal{L}}{\partial x_L}$ and $\frac{\partial \mathcal{L}}{\partial x_L}\left(\frac{\partial}{\partial x_l}\sum_{i=l}^{L-1} \mathcal{F}(x_i, \mathcal{W}_i)\right)$, in which the term $\frac{\partial \mathcal{L}}{\partial x_L}$ is independent of any weight layers. This additive term ensures that the information flow can be easily propagated back from any deeper unit $L$ to any shallower unit $l$. Based on the above analyses, it can be concluded that if we replace ReLU layers after element-wise additions by identity mappings, each ResNet unit will have a direct path to the gradients from the loss function and to the input signal. In other words, the information flow can be directly propagated from any unit to another, both forward and backward for the entire network.

To implement the computations as described in Eqn. (6),

we remove all ReLU layers behind element-wise additions ⊕. In addition, BN is used before each convolutional layer, ReLU is adopted right after BN. This order allows to improve regularization of the network. Dropout (Hinton et al., 2012) with a rate of **0.5** is used to prevent overfitting and located between two convolutional layers. With this architecture, the mapping function $\mathcal{F}(\cdot)$ is executed via a sequence of layers: **BN-ReLU-Convolution-BN-ReLU-Dropout-Convolution** as shown in Figure 8b.

## 4. Experiments

In this section, we experiment the proposed method on three 3D skeleton datasets. We first present the datasets and their evaluation criteria. Some data augmentation techniques that are used for generating more training data are then described. Finally, implementation details of our deep networks are provided.

### *4.1. Datasets and evaluation criteria*

In this work, we evaluate the proposed deep learning framework on MSR Action 3D (Li et al., 2010), KARD (Gaglio et al., 2015) and NTU-RGB+D (Shahroudy et al., 2016). For each dataset, we follow the same evaluation criteria as provided in the original papers. For the interested reader, some public RGB-D datasets for HAR can be found in recent surveys (Zhang et al., 2016; Liu et al., 2017b).

#### *4.1.1. MSR Action 3D dataset*

The MSR Action 3D dataset[1] (Li et al., 2010) consists of 20 different action classes. Each action is performed by 10 subjects for three times. There are 567 skeleton sequences in total. However, 10 sequences are not valid since the skeletons were either missing. Therefore, our experiment was conducted on 557 sequences. For each skeleton frame, the 3D coordinates of 20 joints are provided. The authors of this dataset suggested dividing the whole dataset into three subsets, named **AS1**, **AS2**, and **AS3**. The list of actions for each subset is shown in Table 1. For each subset, we follow the cross-subject evaluation method used by many other authors working with this dataset. More specifically, a half of the dataset (subjects with **ID**s: **1**, **3**, **5**, **7**, **9**) is selected for training and the rest (subjects with **ID**s: **2**, **4**, **6**, **8**, **10**) for test.

| **AS1** | **AS2** | **AS3** |
|---|---|---|
| *Horizontal arm wave* | *High arm wave* | *High throw* |
| *Hammer* | *Hand catch* | *Forward kick* |
| *Forward punch* | *Draw* x | *Side kick* |
| *High throw* | *Draw tick* | *Jogging* |
| *Hand clap* | *Draw circle* | *Tennis swing* |
| *Bend* | *Two hand wave* | *Tennis serve* |
| *Tennis serve* | *Forward kick* | *Golf swing* |
| *Pickup & Throw* | *Side-boxing* | *Pickup & Throw* |

**Table 1. The list of actions in three subsets AS1, AS2, and AS3 of the MSR Action 3D dataset (Li et al., 2010).**

#### *4.1.2. Kinect Activity Recognition Dataset (KARD)*

This dataset[2] (Gaglio et al., 2015) contains 18 actions, performed by 10 subjects and each subject repeated each action three times. KARD provides 540 skeleton sequences in total. Each frame comprises 15 joints. The authors suggested the different evaluation methods on this dataset in which the whole dataset is divided into three subsets as shown in Table 2. For each subset, three experiments have been proposed. Experiment **A** uses one-third of the dataset for training and the rest for test. Meanwhile, experiment **B** uses two-third of the dataset for training and one-third for test. Finally, experiment **C** uses a half of the dataset for training and the remainder for testing.

[1]The MSR Action 3D dataset can be obtained at: `https://www.uow.edu.au/~wanqing/#Datasets`.

[2]The KARD dataset can be obtained at: `https://data.mendeley.com/datasets/k28dtm7tr6/1`.

| Action Set 1 | Action Set 2 | Action Set 3 |
|---|---|---|
| *Horizontal arm wave* | *High arm wave* | *Draw tick* |
| *Two-hand wave* | *Side kick* | *Drink* |
| *Bend* | *Catch cap* | *Sit down* |
| *Phone call* | *Draw tick* | *Phone call* |
| *Stand up* | *Hand clap* | *Take umbrella* |
| *Forward kick* | *Forward kick* | *Toss paper* |
| *Draw X* | *Bend* | *High throw* |
| *Walk* | *Sit down* | *Horiz. arm wave* |

**Table 2. The list of action classes in each subset of the KARD dataset (Gaglio et al., 2015).**

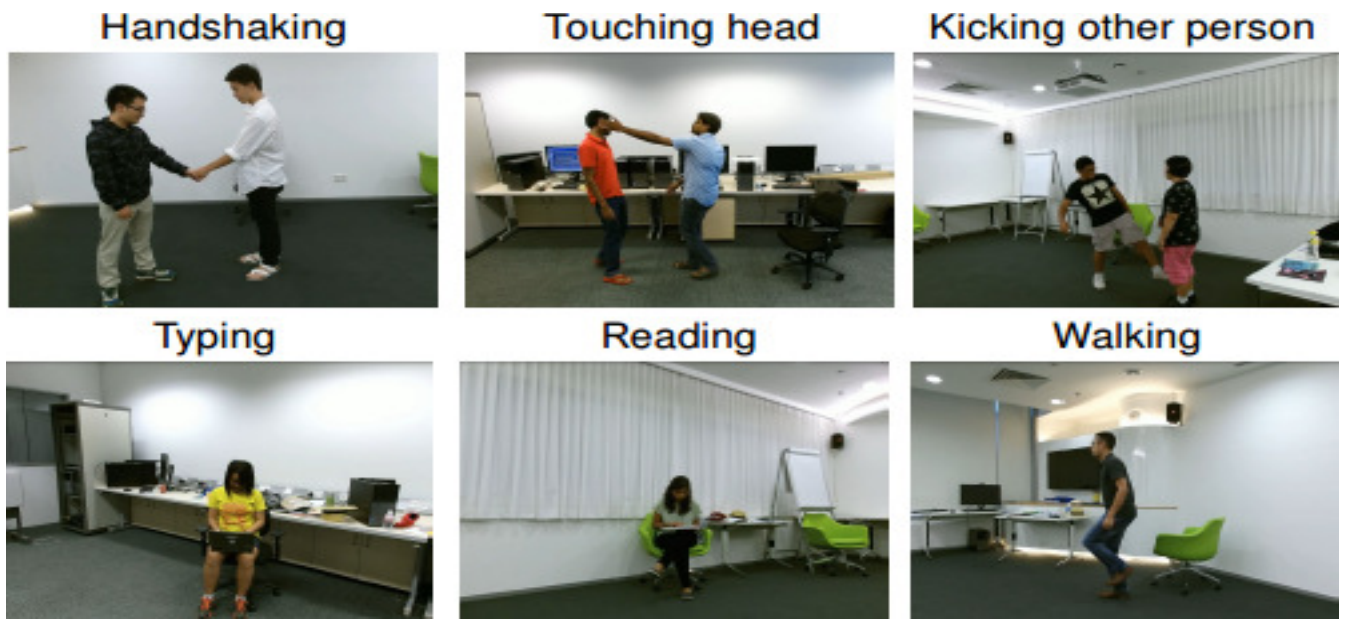

**Fig. 9. Some action classes of the NTU-RGB+D dataset (Shahroudy et al., 2016).**

### 4.1.3. *NTU-RGB+D Action Recognition Dataset*

The NTU-RGB+D[3] (Shahroudy et al., 2016) is a very large-scale dataset. To the best of our knowledge, this is the largest and state-of-the-art RGB-D/skeleton dataset for HAR currently available. It provides more than 56 thousand video samples and 4 million frames, collected from 40 distinct subjects for 60 different action classes. Figure 9 shows some action classes of this dataset. The full list of action classes is provided in **Appendix A**. The 3D skeletal data contains the 3D coordinates of 25 major body joints (Figure 10) provided by Kinect v2 sensor. Therefore, its skeletal data describes more accurately about human movements. The author of this dataset suggested two different evaluation criteria including Cross-Subject and Cross-View. For Cross-Subject evaluation, the sequences performed by 20 subjects with IDs: **1**, **2**, **4**, **5**, **8**, **9**, **13**, **14**, **15**, **16**, **17**, **18**, **19**, **25**, **27**, **28**, **31**, **34**, **35**, and **38** are used for training and the rest sequences are used as testing data. In the Cross-View evaluation, the sequences provided by cameras **2** and **3** are used for training while sequences from camera **1** are used for test.

[3]The NTU-RGB+D dataset can be obtained at: `http://rose1.ntu.edu.sg/Datasets/actionRecognition.asp` with authorization.

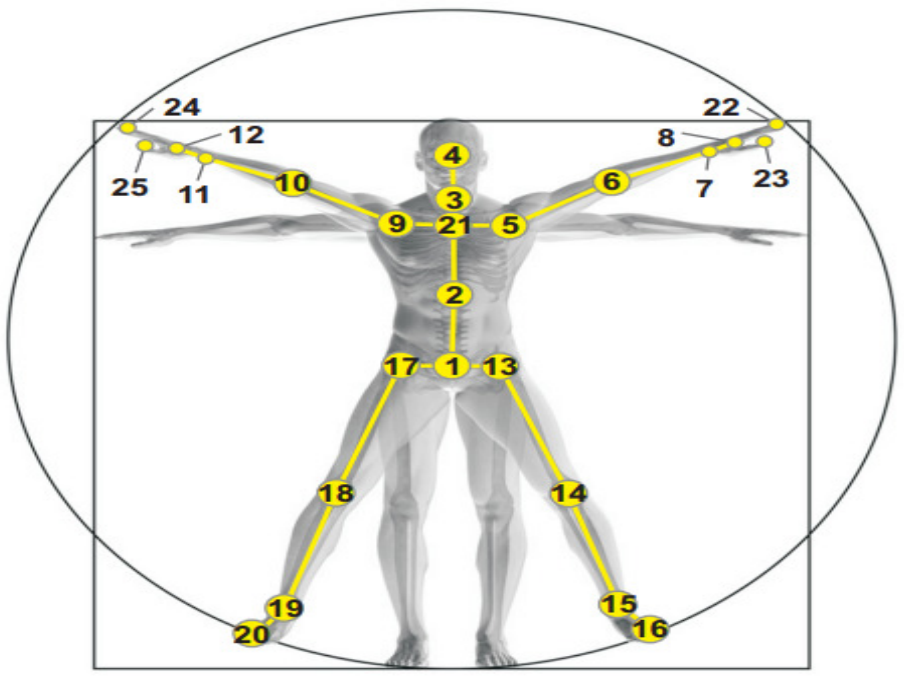

**Fig. 10. Configuration of 25 body joints in each frame of the NTU+RGBD dataset (Shahroudy et al., 2016).**

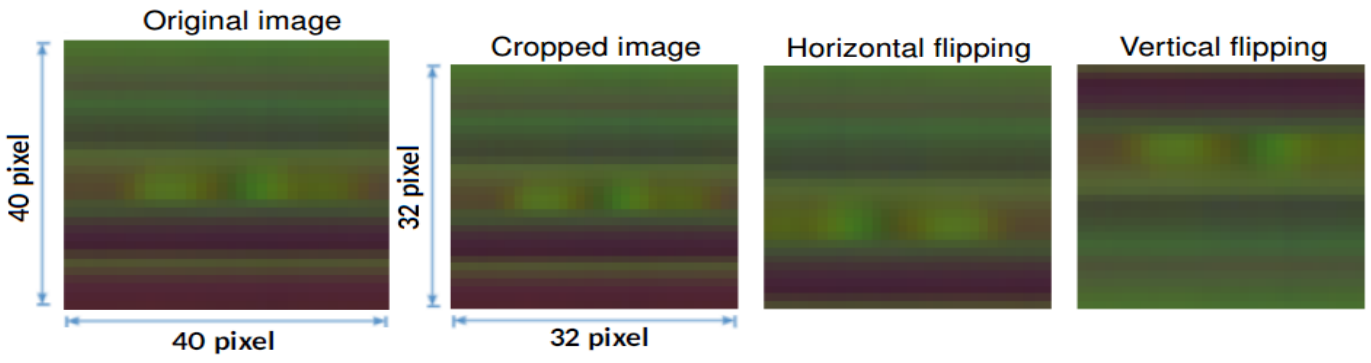

**Fig. 11. Data augmentation applied on MSR Action 3D dataset.**

### 4.2. *Data augmentation*

Very deep neural networks require a lot of data to train. Unfortunately, we have only 557 skeleton sequences on MSR Action 3D dataset (Li et al., 2010) and 540 sequences on KARD (Gaglio et al., 2015). Thus, to prevent overfitting, some data augmentation techniques have been applied. The random cropping, flip horizontally and vertically techniques were used to generate more training samples. Specifically, **8** × cropping has been applied on **40** × **40** images to create **32** × **32** images. Then, their horizontally and vertically flipped images are created. For the NTU-RGB+D dataset (Shahroudy et al., 2016), due to the very large-scale of this dataset, data augmentation techniques were not applied.

### 4.3. *Implementation details*

Different configurations of ResNet with 20-layers, 32-layers, 44-layers, 56-layers, and 110-layers were designed, based on the original Resnet (He et al., 2016) building unit (Figure 8a) and the proposed ResNet building unit (Figure 8b). Totally, we have ten different ResNets. The baseline of the proposed architectures can be found in **Appendix B**. All networks are designed for the acceptable images with the size of $\mathbf{32} \times \mathbf{32}$ pixels as input features and classifying them into **n** categories corresponding to **n** action classes in each dataset. In the experiments, we use a mini-batch of size **128** for 20-layer, 32-layer, 44-layer, and 56-layer networks and a mini-batch of size **64** for 110-layer networks. We initialize the weights randomly and train all networks in an end-to-end manner using **S**tochastic **G**radient **D**escent (**SGD**) algorithm (Bottou, 2010) for **200** epochs from scratch. The learning rate starts from **0.01** for the first 75 epochs, **0.001** for the next 75 epochs and **0.0001** for the remaining 50 epochs. The weight decay is set at **0.0001** and the momentum at **0.9**. In this project, MatConvNet[4] (Vedaldi and Lenc, 2015) is used to implement the solution. Our code, models, and pre-trained models will be shared with the community at: `https://github.com/huyhieupham`.

## 5. Experimental results

This section reports our experimental results. To show the effectiveness of the proposed method, the achieved results are compared with the state-of-the-art methods in literature. All these comparisons are made under the same evaluation criteria.

### *5.1. Results on MSR Action 3D dataset*

The experimental results on MSR Action 3D dataset (Li et al., 2010) are shown in Table 3. We achieved the best classification accuracy with the 44-layer ResNet model which is constructed from the proposed ResNet building unit. Specifically, classification accuracies are **99.9%** on **AS1**, **99.8%** on **AS2**, and **100%** on **AS3**. We obtained a total average accuracy of **99.9%**. Table 5 compares the performance between our best result with the state-of-the-art methods reported on this benchmark. This comparison indicates that the proposed model outperforms many prior works, in which we improved the accuracy rate by **3.4%** compared to the best previous published results. Figure 12a and Figure 12b show the learning curves of all networks on **AS1** subset.

| Model | AS1 | AS2 | AS3 | Aver. |
|---|---|---|---|---|
| Original-ResNet-20 | 99.5% | 98.6% | 99.9% | 99.33% |
| Original-ResNet-32 | 99.5% | 99.1% | 99.9% | 99.50% |
| Original-ResNet-44 | 99.6% | 98.5% | 100% | 99.37% |
| Original-ResNet-56 | 99.3% | 98.4% | 99.5% | 99.07% |
| Original-ResNet-110 | 99.7% | 99.2% | 99.8% | 99.57% |
| Proposed-ResNet-20 | 99.8% | 99.4% | 100% | 99.73% |
| Proposed-ResNet-32 | 99.8% | 99.8% | 100% | 99.87% |
| **Proposed-ResNet-44** | **99.9%** | **99.8%** | **100%** | **99.90%** |
| Proposed-ResNet-56 | 99.9% | 99.1% | 99.6% | 99.53% |
| Proposed-ResNet-110 | 99.9% | 99.5% | 100% | 99.80% |

**Table 3. Recognition accuracy obtained by the proposed method on AS1, AS2, and AS3 subsets of the MSR Action 3D dataset (Li et al., 2010). The best accuracy rates and configuration are highlighted in bold.**

### *5.2. Results on KARD dataset*

The experimental results on KARD dataset (Gaglio et al., 2015) are reported in Table 4. It can be observed the same learning behavior as experiments on MSR Action 3D dataset, in which the best results are achieved by the proposed 44-layer ResNet model. Table 6 provides the accuracy comparison between this model and other approaches on the whole KARD dataset. Based on these comparisons, it can be concluded that our approach outperformed the prior state-of-the-art on this benchmark. Figure 12c and Figure 12d show the learning curves of all networks on **Activity Set 1** subset for **Experiment C**.

### *5.3. Results on NTU RGB-D dataset*

Table 7 shows the experimental results on NTU-RGB+D dataset (Shahroudy et al., 2016). The best network achieved an accuracy of **78.2%** on the Cross-Subject evaluation and **85.6%**

[4]MatconvNet is an open source library for implementing Convolutional Neural Networks (CNNs) in the Matlab environment and can be downloaded at address: `http://www.vlfeat.org/matconvnet/`.

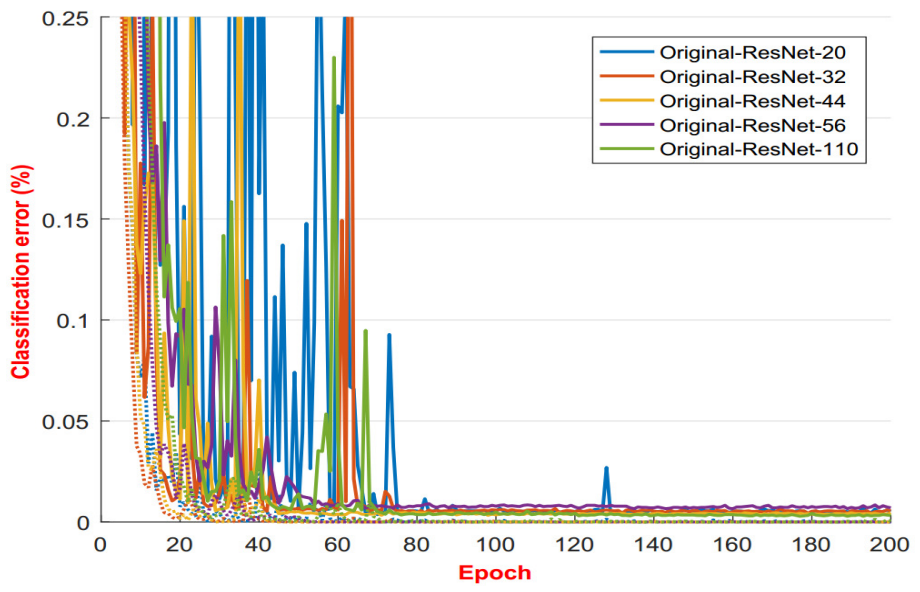

(a) Original ResNets on MSR Action 3D (**AS1 subset**)

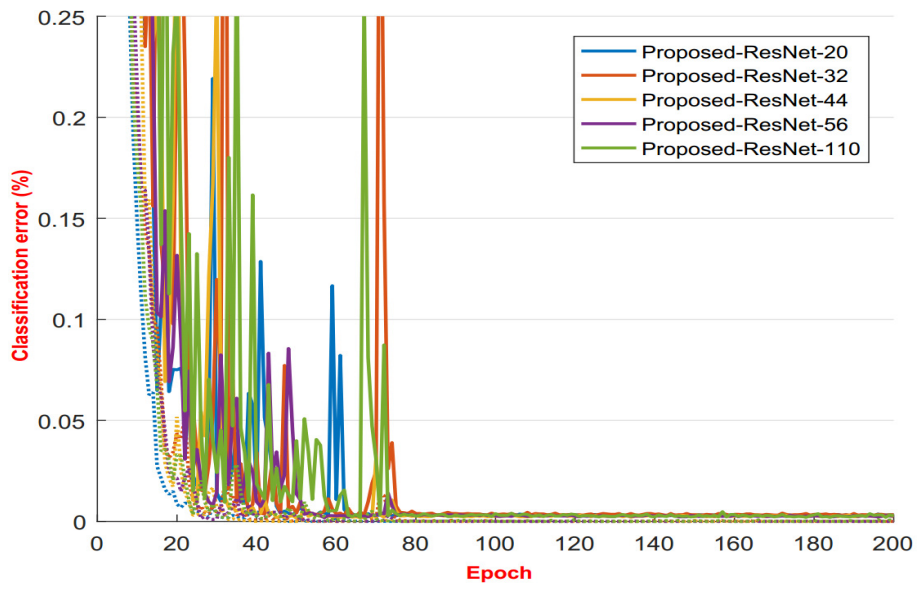

(b) Proposed ResNets on MSR Action 3D (**AS1 subset**)

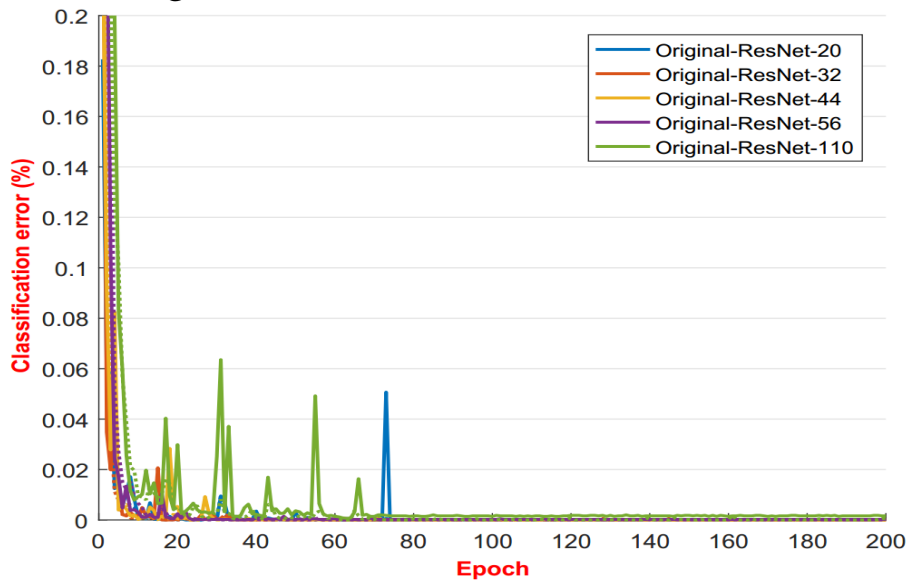

(c) Original ResNets on KARD (**Activity 1**; **Experiment** C)

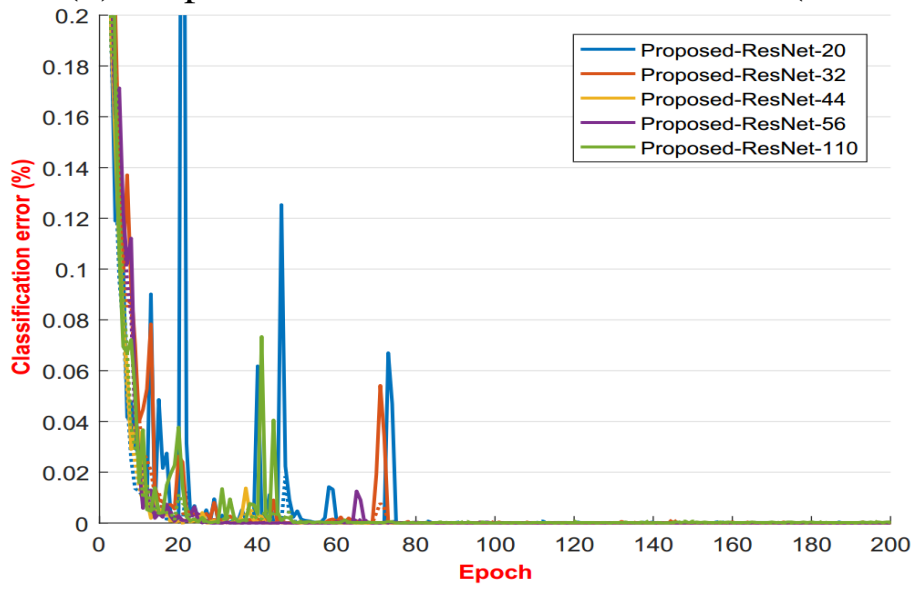

(d) Proposed ResNets on KARD (**Activity 1**; **Experiment C**)

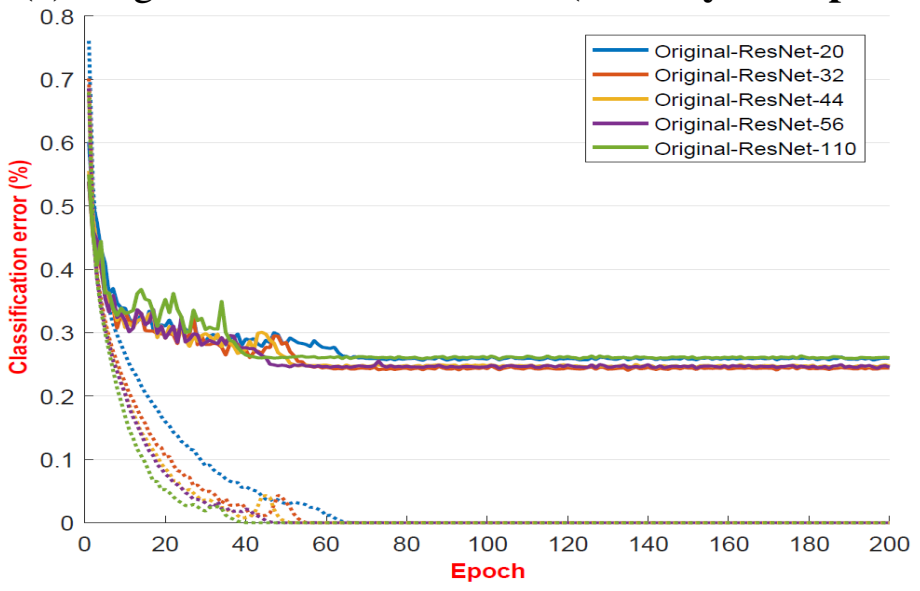

(e) Original ResNets on NTU-RGB+D (**Cross-Subject**)

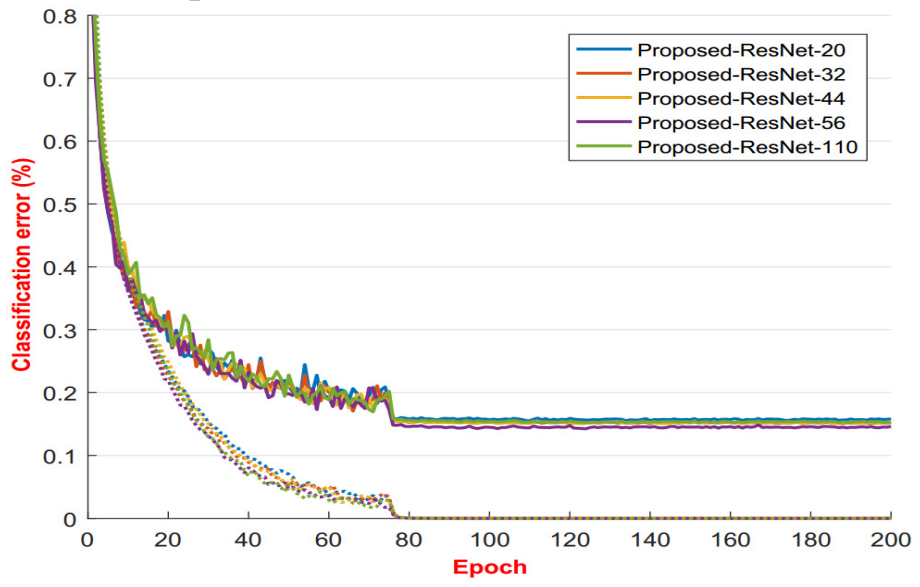

(f) Proposed ResNets on NTU-RGB+D (**Cross-Subject**)

**Fig. 12. Learning curves on MSR Action 3D (Li et al., 2010), KARD (Gaglio et al., 2015) and NTU-RGB+D (Shahroudy et al., 2016) datasets. Dashed lines denote training errors, bold lines denote test errors. We recommend the reader to use a computer and zoom in to see these figures.**

| | **Activity Set 1** | | | **Activity Set 2** | | | **Activity Set 3** | | |
|---|---|---|---|---|---|---|---|---|---|
| **Model** | **Exp. A** | **Exp. B** | **Exp. C** | **Exp. A** | **Exp. B** | **Exp. C** | **Exp. A** | **Exp. B** | **Exp. C** |
| Original-Resnet-20 | 100% | 100% | 100% | 100% | 100% | 100% | 99.8% | 100% | 99.8% |
| Original-ResNet-32 | 100% | 100% | 100% | 100% | 100% | 100% | 99.8% | 99.9% | 99.8% |
| Original-ResNet-44 | 100% | 100% | 100% | 100% | 100% | 100% | 99.7% | 99.7% | 99.7% |
| Original-ResNet-56 | 99.9% | 100% | 100% | 100% | 100% | 99.9% | 99.5% | 99.9% | 99.8% |
| Original-ResNet-110 | 99.8% | 100% | 99.8% | 99.9% | 100% | 99.9% | 99.3% | 100% | 99.7% |
| Proposed-Resnet-20 | 100% | 100% | 100% | 100% | 100% | 100% | 99.8% | 100% | 99.9% |
| Proposed-ResNet-32 | 100% | 100% | 100% | 100% | 100% | 100% | 99.8% | 99.9% | 99.8% |
| **Proposed-ResNet-44** | **100%** | **100%** | **100%** | **100%** | **100%** | **100%** | **99.9%** | **99.9%** | **100%** |
| Proposed-ResNet-56 | 100% | 100% | 100% | 100% | 100% | 100% | 99.7% | 100% | 99.8% |
| Proposed-ResNet-110 | 99.9% | 100% | 99.9% | 100% | 100% | 100% | 99.7% | 100% | 99.8% |

**Table 4. Recognition accuracy obtained by the proposed method on KARD dataset (Gaglio et al., 2015). The best accuracy rates and configuration are highlighted in bold.**

| Method | AS1 | AS2 | AS3 | Aver. |
|---|---|---|---|---|
| (Li et al., 2010) | 72.90% | 71.90% | 79.20% | 74.67% |
| (Vieira et al., 2012) | 84.70% | 81.30% | 88.40% | 84.80% |
| (Xia et al., 2012a) | 87.98% | 85.48% | 63.46% | 78.97% |
| (Chaaraoui et al., 2013) | 92.38% | 86.61% | 96.40% | 91.80% |
| (Chen et al., 2013) | 96.20% | 83.20% | 92.00% | 90.47% |
| (Luo et al., 2013) | 97.20% | 95.50% | 99.10% | 97.26% |
| (Gowayyed et al., 2013) | 92.39% | 90.18% | 91.43% | 91.26% |
| (Hussein et al., 2013) | 88.04% | 89.29% | 94.29% | 90.53% |
| (Qin et al., 2013) | 81.00% | 79.00% | 82.00% | 80.66% |
| (Liang and Zheng, 2013) | 73.70% | 81.50% | 81.60% | 78.93% |
| (Evangelidis et al., 2014) | 88.39% | 86.61% | 94.59% | 89.86% |
| (Ilias et al., 2014) | 91.23% | 90.09% | 99.50% | 93.61% |
| (Gao et al., 2014) | 92.00% | 85.00% | 93.00% | 90.00% |
| (Vieira et al., 2014) | 91.70% | 72.20% | 98.60% | 87.50% |
| (Chen et al., 2015) | 98.10% | 92.00% | 94.60% | 94.90% |
| (Du et al., 2015) | 93.33% | 94.64% | 95.50% | 94.49% |
| (Xu et al., 2015) | 99.10% | 92.90% | 96.40% | 96.10% |
| (Jin et al., 2017) | 99.10% | 92.30% | 98.20% | 96.50% |
| **Our best model** | **99.90%** | **99.80%** | **100%** | **99.90%** |

**Table 5. Comparing our best performance (Proposed-ResNet-44) with other approaches on the MSR Action 3D dataset (Li et al., 2010). All methods use the same experimental protocol.**

| Method | Exp. A | Exp. B | Exp. C | Aver. |
|---|---|---|---|---|
| (Gaglio et al., 2015) | 89.73% | 94.50% | 88.27% | 90.83% |
| (Cippitelli et al., 2016b) | 96.47% | 98.27% | 96.87% | 97.20% |
| (Ling et al., 2016) | 98.90% | 99.60% | 99.43% | 99.31% |
| **Our best model** | **99.97%** | **99.97%** | **100%** | **99.98%** |

**Table 6. Average recognition accuracy of the best proposed model (Proposed-ResNet-44) for experiments A, B and C compared to other approaches on the whole KARD dataset (Gaglio et al., 2015).**

| Model | Cross-Subject | Cross-View |
|---|---|---|
| Original-ResNet-20 | 73.90% | 80.80% |
| Original-ResNet-32 | 75.40% | 81.60% |
| Original-ResNet-44 | 75.20% | 81.50% |
| Original-ResNet-56 | 75.00% | 81.50% |
| Original-ResNet-110 | 73.80% | 80.00% |
| Proposed-ResNet-20 | 76.80% | 83.80% |
| Proposed-ResNet-32 | 76.70% | 84.70% |
| Proposed-ResNet-44 | 77.20% | 84.80% |
| **Proposed-ResNet-56** | **78.20%** | **85.60%** |
| Proposed-ResNet-110 | 78.00% | 84.60% |

**Table 7. Recognition accuracy on NTU-RGB+D dataset (Shahroudy et al., 2016) for Cross-Subject and Cross-View evaluations. The best accuracy rates and configuration are highlighted in bold.**

on the Cross-View. The performance comparison between the proposed method and the state-of-the-art methods on these two evaluations are provided in Table 8 and Table 9. These results showed that our proposed method can deal with very large-scale datasets and outperforms various state-of-the-art approaches for both evaluations. Comparing with the best published result reported by (Li et al., 2017) for the Cross-Subject evaluation, our method significantly surpassed this result by a margin of **+2.0%**. For the Cross-View evaluation, we outperformed the state-of-the-art accuracy in (Kim and Reiter, 2017) by a margin of **+2.5%**. Figure 12e and Figure 12f show the learning curves of all networks in these experiments.

## 6. Discussion

An efficient and effective deep learning framework for HAR should be able to recognize actions with high accuracy and to have the ability to prevent overfitting. Additionally, another important factor is computational efficiency. In this section, these aspects of the proposed method will be evaluated. We also discuss the degradation phenomenon – an important aspect in training very deep learning networks.

| Method (protocol of (Shahroudy et al., 2016)) | Cross-Subject |
|---|---|
| HON4D (Oreifej and Liu, 2013) | 30.56% |
| Super Normal Vector (Yang and Tian, 2014) | 31.82% |
| HOG$^2$ (Ohn-Bar and Trivedi, 2013) | 32.24% |
| Skeletal Quads (Evangelidis et al., 2014) | 38.62% |
| Shuffle and Learn (Misra et al., 2016) | 47.50% |
| Key poses + SVM (Cippitelli et al., 2016a) | 48.90% |
| Lie Group (Vemulapalli et al., 2014) | 50.08% |
| HBRNN-L (Du et al., 2015) | 59.07% |
| FTP Dynamic Skeletons (Hu et al., 2015) | 60.23% |
| P-LSTM (Shahroudy et al., 2016) | 62.93% |
| RNN Encoder-Decoder (Luo et al., 2017) | 66.20% |
| ST-LSTM (Liu et al., 2016) | 69.20% |
| STA-LSTM (Song et al., 2017) | 73.40% |
| Res-TCN (Kim and Reiter, 2017) | 74.30% |
| DSSCA - SSLM (Shahroudy et al., 2017) | 74.86% |
| Joint Distance Maps + CNN (Li et al., 2017) | 76.20% |
| **Our best model (Proposed-ResNet-56)** | **78.20%** |

**Table 8. Performance comparison of our proposed ResNet models with the state-of-the-art methods on the Cross-Subject evaluation criteria of NTU-RGB+D dataset (Shahroudy et al., 2016).**

### 6.1. *Classification accuracy*

In Section 5, we evaluated the proposed learning framework on three well-known benchmark datasets. We demonstrate empirically that our method outperforms many previous studies on all these datasets under the same experimental protocols. The improvements on each benchmark are shown in Table 10. It is clear that in terms of accuracy, our learning model is effective for solving the problems of HAR.

### 6.2. *Overfitting issues and degradation phenomenon*

Considering the accuracies obtained by our proposed ResNet architecture and comparing them to the results obtained by the original ResNet architecture, we observed that our proposed networks are able to reduce the effects of the degradation phenomenon for both training and test phases. *E.g.*, the proposed 56-layer networks achieved better results than 20-layer, 32-layer, and 44-layer networks on NTU-RGB+D dataset. Meanwhile, the original ResNet with 32-layer is the best network on this benchmark. The same learning behaviors are found in experiments on the MSR Action 3D and KARD datasets (Table 11). It should be noted that degradation phenomena depend considerably on the size of datasets[5]. This is the reason why the 110-layer network got higher errors than several other networks. The difference between training error and test error on the learning curves shows the ability of overfitting prevention.

| Method (protocol of (Shahroudy et al., 2016)) | Cross-View |
|---|---|
| HON4D (Oreifej and Liu, 2013) | 7.26% |
| Super Normal Vector (Yang and Tian, 2014) | 13.61% |
| HOG$^2$ (Ohn-Bar and Trivedi, 2013) | 22.27% |
| Skeletal Quads (Evangelidis et al., 2014) | 41.36% |
| Lie Group (Vemulapalli et al., 2014) | 52.76% |
| Key poses + SVM (Cippitelli et al., 2016a) | 57.70% |
| HBRNN-L (Du et al., 2015) | 63.97% |
| FTP Dynamic Skeletons (Hu et al., 2015) | 65.22% |
| P-LSTM (Shahroudy et al., 2016) | 70.27% |
| ST-LSTM (Liu et al., 2016) | 77.7% |
| STA-LSTM (Song et al., 2017) | 81.2% |
| Joint Distance Maps + CNN (Li et al., 2017) | 82.3% |
| Res-TCN (Kim and Reiter, 2017) | 83.1% |
| **Our best model (Proposed-ResNet-56)** | **85.60%** |

**Table 9. Performance comparison of our proposed ResNet models with the state-of-the-art methods on the Cross-View evaluation criteria of NTU-RGB+D dataset (Shahroudy et al., 2016).**

| | MRS 3D (overall) | KARD (overall) | NTU-RGB+D Cross-Subject | NTU-RGB+D Cross-View |
|---|---|---|---|---|
| Prior works | 96.50% | 99.31% | 76.20% | 83.10% |
| Our results | 99.90% | 99.98% | 78.20% | 85.60% |
| Improvements | **3.40%** | **0.67%** | **2.00%** | **2.50%** |

**Table 10. The best of our results compared to the best prior results on MSR Action 3D, KARD, and NTU-RGB+D datasets.**

[5]Personal communication with H. Zang from the Rutgers University, USA and Amazon AI.

| # Network layers | MSR Action 3D | KARD | NTU-RGB+D |
|---|---|---|---|
| 110 | | | |
| 56 | | | ✗ |
| 44 | ✗ | ✗ | |
| 32 | ✓ | | ✓ |
| 20 | | ✓ | |

**Table 11. Relationship between the number of layers and its performance on three benchmarks. ✓ denotes the best network based on the original ResNet architecture and ✗ denotes the best network based on the proposed ResNet architecture.**

Our experimental results on three action benchmarks provided that the proposed ResNet architectures are capable of reducing overfitting in comparison with the original architecture. We believe this result comes from the combination between the use of BN (Ioffe and Szegedy, 2015b) before convolutional layers and Dropout (Hinton et al., 2012) in each ResNet unit.

### 6.3. *Effect of image resizing methods on recognition result*

D-CNNs work with fixed size tensors. Thus, before feeding image-based representations to ResNets, all these images were resized to a fixed size of $\mathbf{32} \times \mathbf{32}$ pixels. The resizing step may lead to the change in the accuracy rate. To identify the effects of different resizing methods on the recognition performance of the proposed model, we conducted an additional experiment on the MSR Action 3D/AS1 dataset (Li et al., 2010) with Proposed-ResNet-20 network. In this experiment, two different resizing methods, including Nearest-Neighbor interpolation and Bicubic interpolation were used for resizing image-based representations before feeding to deep networks. Experimental results indicate that the difference between the accuracy rates is very small ($\Delta$ = **0.3**%; see Figure 13).

### 6.4. *Effect of joint order on recognition result*

In our study, each skeleton was divided into five parts and concatenated in a certain order in order to keep the local motion characteristics and to generate discriminative features in image-based representations. To clarify the effect of order of joints in skeletons, we have tried to remove the step of rearranging joints in our implementation and perform experiments with the order of joints provided by the Kinect SDK. We observed a dramatically decrease in the recognition accuracy ($\Delta$ = **9.0**%) as shown in Figure 14.

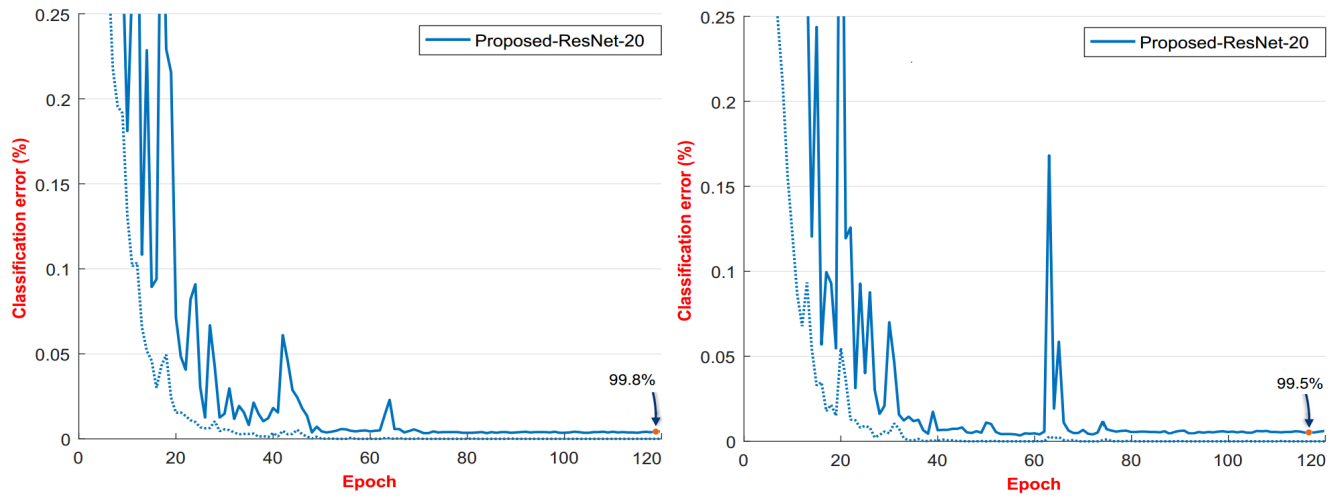

(a) (b)

**Fig. 13. Training and test errors (%) by the Proposed-ResNet-20 network on the MSR Action 3D/AS1 dataset (Li et al., 2010): (a) resizing images using Bicubic interpolation; (b) resizing images using Nearest-neighbor interpolation.**

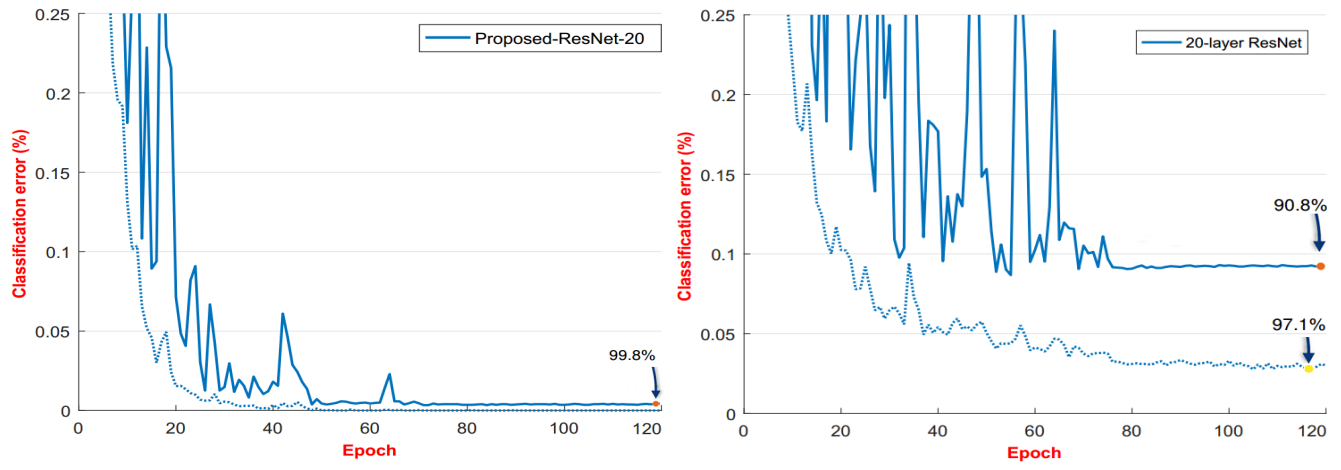

(a) (b)

**Fig. 14. Training and test errors (%) by the Proposed-ResNet-20 network on the MSR Action 3D/AS1 dataset (Li et al., 2010): (a) rearranging skeletons according to human body physical structure; (b) using the joints order provided by the Kinect SDK without rearranging skeletons.**

### 6.5. *Computational efficiency*

We take the Cross-View evaluation criterion of the NTU-RGB+D dataset (Shahroudy et al., 2016) and the Proposed-ResNet-56 network to illustrate the computational efficiency of our learning framework. As shown in Figure 15, the proposed method has main components, including Stage 1 the encoding process from skeletons to RGB images, Stage 2 the supervised training stage, and Stage 3 the prediction stage. With the implementation in Matlab using MatConvNet toolbox (Vedaldi and Lenc, 2015) on a single NVIDIA GeForce

GTX 1080 Ti GPU system[6], without parallel processing, we take **7.83** $\times$ **10**$^{-3}$s per skeleton sequence during training. After about **80** epochs, our network starts converging with an accuracy around **85%**. While the prediction time, including the time for encoding skeletons into RGB images and classification by pre-trained ResNet, takes **0.128**s per skeleton sequence. This speed is fast enough to respond to many real-time applications. Our method is therefore more practical for large-scale problems and real-time applications.

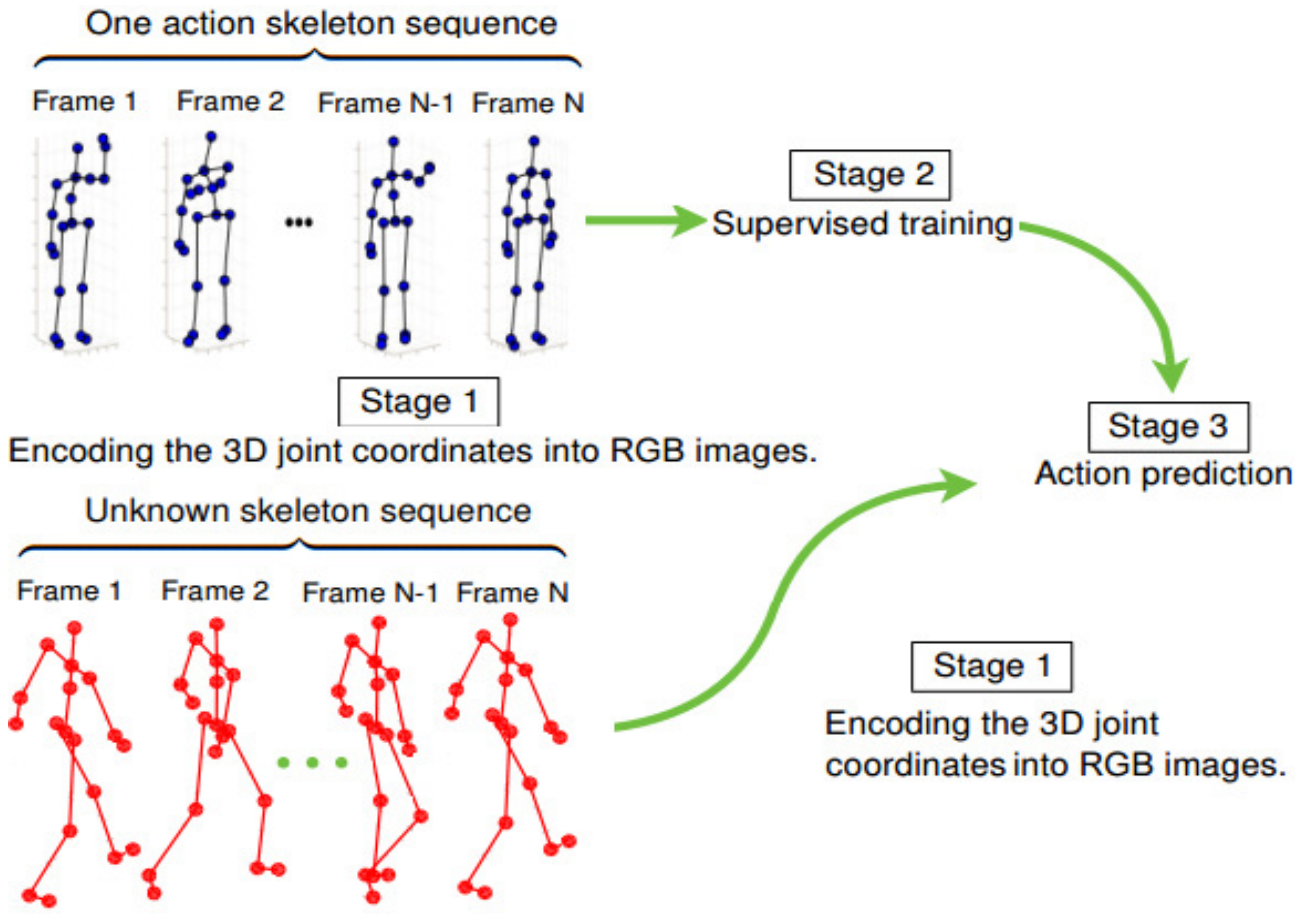

**Fig. 15. Three main phases in the proposed method.**

| Component | Average processing time |
|---|---|
| Stage 1 | **7.83** $\times$ **10**$^{-3}$s per sequence |
| Stage 2 | **1.27** $\times$ **10**$^{-3}$s per sequence |
| Stage 3 | **0.128**s per sequence |

**Table 12. Execution time of each component of our method.**

## 7. Conclusion and future work

We have presented a novel deep learning framework based on ResNets for human action recognition with skeletal data. The idea is to combine two important factors, *i.e.*, a good spatio-temporal representation of 3D motion and a powerful deep learning model. By encoding skeleton sequences into RGB images and proposing a novel ResNet architecture for learning human action from these images, higher levels of performance have been achieved. We show that the approach is effective for recognizing actions as we achieved state-of-the-art performance on three well-established datasets, while requiring less computation resource. We are currently extending the skeleton encoding method in which the Euclidean distance and the orientation relationship between joints are exploited. In addition, to achieve a better feature learning and classification framework, we aim to design and train some new and potential D-CNN architectures based on the idea of ResNet (He et al., 2016) such as Inception-ResNet-v2 (Szegedy et al., 2017) and Densely Connected Convolutional Networks (DenseNet) (Huang et al., 2016). The preliminary results are encouraging. We hope that our study opens a new door for the research community on exploiting the potentials of very deep networks for human action recognition.

[6]For more information about the specifications of this GPU, please refer to: https://www.nvidia.com/en-us/geforce/products/10series/geforce-gtx-1080-ti/.

## Acknowledgments

This research was carried out at the Cerema Research Center (CEREMA) and Toulouse Institute of Computer Science Research (IRIT), France. Sergio A Velastin is grateful for funding received from the Universidad Carlos III de Madrid, the European Union's Seventh Framework Programme for research, technological development and demonstration under grant agreement n 600371, el Ministerio de Economía, Industria y Competitividad (COFUND2013-51509) el Ministerio de Educación, Cultura y Deporte (CEI-15-17) and Banco Santander.

## Appendix

### A. List of action classes from the NTU-RGB+D dataset

Below is the list of the action classes provided by the NTU-RGB+D dataset (Shahroudy et al., 2016), it contains 60 different actions captured by Kinect v2 sensors:

*Drinking, eating, brushing teeth, brushing hair, dropping, picking up, throwing, sitting down, standing up, clapping, reading, writing, tearing up paper, wearing jacket, taking off jacket, wearing a shoe, taking off a shoe, wearing on glasses, taking off glasses, puting on a hat/cap, taking off a hat/cap, cheering up, hand waving, kicking something, reaching into self pocket, hopping, jumping up, making/answering a phone call, playing with phone, typing, pointing to something, taking selfie, checking time, rubbing two hands together, bowing, shaking head, wiping face, saluting, putting palms together, crossing hands in front. sneezing/coughing, staggering, falling down, touching head, touching chest, touching back, touching neck, vomiting, fanning self. punching/slapping other person, kicking other person, pushing other person, patting others back, pointing to the other person, hugging, giving something to other person, touching other persons pocket, handshaking, walking towards each other, and walking apart from each other.*

### B. Network structures

This section describes the network architectures in detail. To build 20-layer, 32-layer, 44-layer, 56-layer, and 110-layer networks, we stack the proposed ResNet building units as following:

| **Baseline 20-layer ResNet architecture** |
|---|
| 3x3 Conv., 16 filters, BN, ReLU |
| Residual unit: BN-ReLU-Conv.-BN-ReLU-Dropout-Conv.,16 filters |
| Residual unit: BN-ReLU-Conv.-BN-ReLU-Dropout-Conv.,16 filters |
| Residual unit: BN-ReLU-Conv.-BN-ReLU-Dropout-Conv.,16 filters |
| Residual unit: BN-ReLU-Conv.-BN-ReLU-Dropout-Conv.,32 filters |
| Residual unit: BN-ReLU-Conv.-BN-ReLU-Dropout-Conv.,32 filters |
| Residual unit: BN-ReLU-Conv.-BN-ReLU-Dropout-Conv.,32 filters |
| Residual unit: BN-ReLU-Conv.-BN-ReLU-Dropout-Conv.,64 filters |
| Residual unit: BN-ReLU-Conv.-BN-ReLU-Dropout-Conv.,64 filters |
| Residual unit: BN-ReLU-Conv.-BN-ReLU-Dropout-Conv.,64 filters |
| Global mean pooling |
| FC layer with **n** units where **n** is equal the number of action class. |
| Softmax layer |

| **Baseline 32-layer ResNet architecture** |
|---|
| 3x3 Conv., 16 filters, BN, ReLU |
| Residual unit: BN-ReLU-Conv.-BN-ReLU-Dropout-Conv.,16 filters |
| Residual unit: BN-ReLU-Conv.-BN-ReLU-Dropout-Conv.,16 filters |
| Residual unit: BN-ReLU-Conv.-BN-ReLU-Dropout-Conv.,16 filters |
| Residual unit: BN-ReLU-Conv.-BN-ReLU-Dropout-Conv.,16 filters |
| Residual unit: BN-ReLU-Conv.-BN-ReLU-Dropout-Conv.,16 filters |
| Residual unit: BN-ReLU-Conv.-BN-ReLU-Dropout-Conv.,32 filters |
| Residual unit: BN-ReLU-Conv.-BN-ReLU-Dropout-Conv.,32 filters |
| Residual unit: BN-ReLU-Conv.-BN-ReLU-Dropout-Conv.,32 filters |
| Residual unit: BN-ReLU-Conv.-BN-ReLU-Dropout-Conv.,32 filters |
| Residual unit: BN-ReLU-Conv.-BN-ReLU-Dropout-Conv.,32 filters |
| Residual unit: BN-ReLU-Conv.-BN-ReLU-Dropout-Conv.,64 filters |
| Residual unit: BN-ReLU-Conv.-BN-ReLU-Dropout-Conv.,64 filters |
| Residual unit: BN-ReLU-Conv.-BN-ReLU-Dropout-Conv.,64 filters |
| Residual unit: BN-ReLU-Conv.-BN-ReLU-Dropout-Conv.,64 filters |
| Residual block: BN-ReLU-Conv.-BN-ReLU-Dropout-Conv.,64 filters |
| Global mean pooling |
| FC layer with **n** units where **n** is equal the number of action class. |
| Softmax layer |

| **Baseline 44-layer ResNet architecture** |
|---|
| 3x3 Conv., 16 filters, BN, ReLU |
| Residual unit: BN-ReLU-Conv.-BN-ReLU-Dropout-Conv.,16 filters |
| Residual unit: BN-ReLU-Conv.-BN-ReLU-Dropout-Conv.,16 filters |
| Residual unit: BN-ReLU-Conv.-BN-ReLU-Dropout-Conv.,16 filters |
| Residual unit: BN-ReLU-Conv.-BN-ReLU-Dropout-Conv.,16 filters |
| Residual unit: BN-ReLU-Conv.-BN-ReLU-Dropout-Conv.,16 filters |
| Residual unit: BN-ReLU-Conv.-BN-ReLU-Dropout-Conv.,16 filters |
| Residual unit: BN-ReLU-Conv.-BN-ReLU-Dropout-Conv.,16 filters |

Residual unit: BN-ReLU-Conv.-BN-ReLU-Dropout-Conv.,32 filters

Residual unit: BN-ReLU-Conv.-BN-ReLU-Dropout-Conv.,32 filters

Residual unit: BN-ReLU-Conv.-BN-ReLU-Dropout-Conv.,32 filters

Residual unit: BN-ReLU-Conv.-BN-ReLU-Dropout-Conv.,32 filters

Residual unit: BN-ReLU-Conv.-BN-ReLU-Dropout-Conv.,32 filters

Residual unit: BN-ReLU-Conv.-BN-ReLU-Dropout-Conv.,32 filters

Residual unit: BN-ReLU-Conv.-BN-ReLU-Dropout-Conv.,32 filters

Residual unit: BN-ReLU-Conv.-BN-ReLU-Dropout-Conv.,64 filters

Residual unit: BN-ReLU-Conv.-BN-ReLU-Dropout-Conv.,64 filters

Residual unit: BN-ReLU-Conv.-BN-ReLU-Dropout-Conv.,64 filters

Residual unit: BN-ReLU-Conv.-BN-ReLU-Dropout-Conv.,64 filters

Residual unit: BN-ReLU-Conv.-BN-ReLU-Dropout-Conv.,64 filters

Residual unit: BN-ReLU-Conv.-BN-ReLU-Dropout-Conv.,64 filters

Residual unit: BN-ReLU-Conv.-BN-ReLU-Dropout-Conv.,64 filters

Global mean pooling

FC layer with **n** units, where **n** is equal the number of action class.

Softmax layer

**Baseline 56-layer ResNet architecture**

3x3 Conv., 16 filters, BN, ReLU

Residual unit: BN-ReLU-Conv.-BN-ReLU-Dropout-Conv.,16 filters

Residual unit: BN-ReLU-Conv.-BN-ReLU-Dropout-Conv.,16 filters

Residual unit: BN-ReLU-Conv.-BN-ReLU-Dropout-Conv.,16 filters

Residual unit: BN-ReLU-Conv.-BN-ReLU-Dropout-Conv.,16 filters

Residual unit: BN-ReLU-Conv.-BN-ReLU-Dropout-Conv.,16 filters

Residual unit: BN-ReLU-Conv.-BN-ReLU-Dropout-Conv.,16 filters

Residual unit: BN-ReLU-Conv.-BN-ReLU-Dropout-Conv.,16 filters

Residual unit: BN-ReLU-Conv.-BN-ReLU-Dropout-Conv.,16 filters

Residual unit: BN-ReLU-Conv.-BN-ReLU-Dropout-Conv.,16 filters

Residual unit: BN-ReLU-Conv.-BN-ReLU-Dropout-Conv.,32 filters

Residual unit: BN-ReLU-Conv.-BN-ReLU-Dropout-Conv.,32 filters

Residual unit: BN-ReLU-Conv.-BN-ReLU-Dropout-Conv.,32 filters

Residual unit: BN-ReLU-Conv.-BN-ReLU-Dropout-Conv.,32 filters

Residual unit: BN-ReLU-Conv.-BN-ReLU-Dropout-Conv.,32 filters

Residual unit: BN-ReLU-Conv.-BN-ReLU-Dropout-Conv.,32 filters

Residual unit: BN-ReLU-Conv.-BN-ReLU-Dropout-Conv.,32 filters

Residual unit: BN-ReLU-Conv.-BN-ReLU-Dropout-Conv.,32 filters

Residual unit: BN-ReLU-Conv.-BN-ReLU-Dropout-Conv.,32 filters

Residual unit: BN-ReLU-Conv.-BN-ReLU-Dropout-Conv.,64 filters

Residual unit: BN-ReLU-Conv.-BN-ReLU-Dropout-Conv.,64 filters

Residual unit: BN-ReLU-Conv.-BN-ReLU-Dropout-Conv.,64 filters

Residual unit: BN-ReLU-Conv.-BN-ReLU-Dropout-Conv.,64 filters

Residual unit: BN-ReLU-Conv.-BN-ReLU-Dropout-Conv.,64 filters

Residual unit: BN-ReLU-Conv.-BN-ReLU-Dropout-Conv.,64 filters

Residual unit: BN-ReLU-Conv.-BN-ReLU-Dropout-Conv.,64 filters

Residual unit: BN-ReLU-Conv.-BN-ReLU-Dropout-Conv.,64 filters

Residual unit: BN-ReLU-Conv.-BN-ReLU-Dropout-Conv.,64 filters

Global mean pooling

FC layer with **n** units, where **n** is equal the number of action class.

Softmax layer

**Baseline 110-layer ResNet architecture**

3x3 Conv., 16 filters, BN, ReLU

Residual unit: BN-ReLU-Conv.-BN-ReLU-Dropout-Conv.,16 filters

Residual unit: BN-ReLU-Conv.-BN-ReLU-Dropout-Conv.,16 filters

Residual unit: BN-ReLU-Conv.-BN-ReLU-Dropout-Conv.,16 filters

Residual unit: BN-ReLU-Conv.-BN-ReLU-Dropout-Conv.,16 filters

Residual unit: BN-ReLU-Conv.-BN-ReLU-Dropout-Conv.,16 filters

Residual unit: BN-ReLU-Conv.-BN-ReLU-Dropout-Conv.,16 filters

Residual unit: BN-ReLU-Conv.-BN-ReLU-Dropout-Conv.,16 filters

Residual unit: BN-ReLU-Conv.-BN-ReLU-Dropout-Conv.,16 filters

Residual unit: BN-ReLU-Conv.-BN-ReLU-Dropout-Conv.,16 filters

Residual unit: BN-ReLU-Conv.-BN-ReLU-Dropout-Conv.,16 filters

Residual unit: BN-ReLU-Conv.-BN-ReLU-Dropout-Conv.,16 filters

Residual unit: BN-ReLU-Conv.-BN-ReLU-Dropout-Conv.,16 filters

Residual unit: BN-ReLU-Conv.-BN-ReLU-Dropout-Conv.,16 filters

Residual unit: BN-ReLU-Conv.-BN-ReLU-Dropout-Conv.,16 filters

Residual unit: BN-ReLU-Conv.-BN-ReLU-Dropout-Conv.,16 filters

Residual unit: BN-ReLU-Conv.-BN-ReLU-Dropout-Conv.,16 filters

Residual unit: BN-ReLU-Conv.-BN-ReLU-Dropout-Conv.,16 filters

Residual unit: BN-ReLU-Conv.-BN-ReLU-Dropout-Conv.,16 filters

Residual unit: BN-ReLU-Conv.-BN-ReLU-Dropout-Conv.,32 filters

Residual unit: BN-ReLU-Conv.-BN-ReLU-Dropout-Conv.,32 filters

Residual unit: BN-ReLU-Conv.-BN-ReLU-Dropout-Conv.,32 filters

Residual unit: BN-ReLU-Conv.-BN-ReLU-Dropout-Conv.,32 filters

Residual unit: BN-ReLU-Conv.-BN-ReLU-Dropout-Conv.,32 filters

| |
|---|
| Residual unit: BN-ReLU-Conv.-BN-ReLU-Dropout-Conv.,32 filters |
| Residual unit: BN-ReLU-Conv.-BN-ReLU-Dropout-Conv.,32 filters |
| Residual unit: BN-ReLU-Conv.-BN-ReLU-Dropout-Conv.,32 filters |
| Residual unit: BN-ReLU-Conv.-BN-ReLU-Dropout-Conv.,32 filters |
| Residual unit: BN-ReLU-Conv.-BN-ReLU-Dropout-Conv.,32 filters |
| Residual unit: BN-ReLU-Conv.-BN-ReLU-Dropout-Conv.,32 filters |
| Residual unit: BN-ReLU-Conv.-BN-ReLU-Dropout-Conv.,32 filters |
| Residual unit: BN-ReLU-Conv.-BN-ReLU-Dropout-Conv.,32 filters |
| Residual unit: BN-ReLU-Conv.-BN-ReLU-Dropout-Conv.,32 filters |
| Residual unit: BN-ReLU-Conv.-BN-ReLU-Dropout-Conv.,32 filters |
| Residual unit: BN-ReLU-Conv.-BN-ReLU-Dropout-Conv.,32 filters |
| Residual unit: BN-ReLU-Conv.-BN-ReLU-Dropout-Conv.,32 filters |
| Residual unit: BN-ReLU-Conv.-BN-ReLU-Dropout-Conv.,32 filters |
| Residual unit: BN-ReLU-Conv.-BN-ReLU-Dropout-Conv.,64 filters |
| Residual unit: BN-ReLU-Conv.-BN-ReLU-Dropout-Conv.,64 filters |
| Residual unit: BN-ReLU-Conv.-BN-ReLU-Dropout-Conv.,64 filters |
| Residual unit: BN-ReLU-Conv.-BN-ReLU-Dropout-Conv.,64 filters |
| Residual unit: BN-ReLU-Conv.-BN-ReLU-Dropout-Conv.,64 filters |
| Residual unit: BN-ReLU-Conv.-BN-ReLU-Dropout-Conv.,64 filters |
| Residual unit: BN-ReLU-Conv.-BN-ReLU-Dropout-Conv.,64 filters |
| Residual unit: BN-ReLU-Conv.-BN-ReLU-Dropout-Conv.,64 filters |
| Residual unit: BN-ReLU-Conv.-BN-ReLU-Dropout-Conv.,64 filters |
| Residual unit: BN-ReLU-Conv.-BN-ReLU-Dropout-Conv.,64 filters |
| Residual unit: BN-ReLU-Conv.-BN-ReLU-Dropout-Conv.,64 filters |
| Residual unit: BN-ReLU-Conv.-BN-ReLU-Dropout-Conv.,64 filters |
| Residual unit: BN-ReLU-Conv.-BN-ReLU-Dropout-Conv.,64 filters |
| Residual unit: BN-ReLU-Conv.-BN-ReLU-Dropout-Conv.,64 filters |
| Residual unit: BN-ReLU-Conv.-BN-ReLU-Dropout-Conv.,64 filters |
| Residual unit: BN-ReLU-Conv.-BN-ReLU-Dropout-Conv.,64 filters |
| Residual unit: BN-ReLU-Conv.-BN-ReLU-Dropout-Conv.,64 filters |
| Residual unit: BN-ReLU-Conv.-BN-ReLU-Dropout-Conv.,64 filters |
| Global mean pooling |
| FC layer with **n** units, where **n** is equal the number of action class. |
| Softmax layer |